\documentclass[runningheads]{llncs}
\usepackage{graphicx}
\usepackage{float}
\usepackage{amsmath,amssymb,amsfonts}
\usepackage{caption}
\usepackage{subcaption}
\begin{document}
\title{TF-Net: Deep Learning Empowered \textbf{Tiny Feature Network} for Night-time UAV Detection \thanks{This work was supported by the Higher Education Commission (HEC), Pakistan under the NRPU 2021 Grant 15687.
}}
%
%
\author{Maham Misbah\inst{1}\and Misha Urooj Khan\inst{1} \and
Zhaohui Yang\inst{2}\and Zeeshan Kaleem \inst{1}}
\authorrunning{Misbah et al.}
%
\institute{Department of Electrical and Computer Engineering, COMSATS University Islamabad, Wah Campus, Pakistan\\
\email{zeeshankaleem@gmail.com}\\
\and
College of Information Science and Electronic Engineering, Zhejiang University, Hangzhou 310027, China\\
\email{yang\_zhaohui@zju.edu.cn}}
\maketitle              
\begin{abstract}
Technological advancements have normalized the usage of unmanned aerial vehicles (UAVs) in every sector, spanning from military to commercial but they also pose serious security concerns due to their enhanced functionalities and easy access to private and highly secured areas. Several instances related to UAVs have raised security concerns, leading to UAV detection research studies. Visual techniques are widely adopted for UAV detection, but they perform poorly at night, in complex backgrounds, and in adverse weather conditions. Therefore, a robust night vision-based drone detection system is required to that could efficiently tackle this problem. Infrared cameras are increasingly used for nighttime surveillance due to their wide applications in night vision equipment. This paper uses a deep learning-based TinyFeatureNet (TF-Net), which is an improved version of YOLOv5s, to accurately detect UAVs during the night using infrared (IR) images. In the proposed TF-Net, we introduce architectural changes in the neck and backbone of the YOLOv5s. We also simulated four different YOLOv5 models (s,m,n,l) and proposed TF-Net for a fair comparison. The results showed better performance for the proposed TF-Net in terms of precision, IoU, GFLOPS, model size, and FPS compared to the YOLOv5s. TF-Net yielded the best results with 95.7\% precision, 84\% mAp, and 44.8\% $IoU$.

\end{abstract}
\section{Introduction}
Unmanned aerial vehicles (UAVs) are increasingly becoming popular owing to the technological advancements and increased use cases. Initially, drones were only used for military purposes, but now they are used in many other fields  \cite{1}. Currently, the commercial and military industries that rely on UAVs are expanding rapidly, attracting investors who are driving improvements in UAV technology and continuously trying to reduce the size, weight, and prices. Drones started off with bigger size, but as technology advanced, they became smaller and more intelligent \cite{22}. Consequently, these smaller drones have prompted security concerns as they can easily reach any private area \cite{24} \cite{25}.

A lot of research has been conducted to date for the detection of UAVs using visual \cite{5}, radio frequency (RF) \cite{20}, acoustic \cite{24}, thermal \cite{14} and radar methods \cite{18}. But many of them show poor performance during night and in extreme weather conditions. Therefore it is crucial to have a night vision-based drone detection system \cite{15}. The use of infrared cameras for nighttime monitoring has become quite common \cite{23} \cite{18} \cite{14} due to the use of night vision technology in surveillance and protection. The military, law enforcement, and the general public, all sectors have applications where it could be deployed and used effectively. The difficult challenge of overcoming UAV detection at night time could be overcome by the use high-priced and high resolution thermal imaging devices. These high-priced devices have also some limitations like inefficiency in low-light conditions, such as complete darkness, or poor atmospheric conditions when the captured image does not accurately reflect the true things being seen \cite{1}.

The worldwide terrorism problem has expanded this domain into the areas of individual safety, border control, national security, and military surveillance. That's why the development of IR based object detection and classification are among the fastest topics being worked on \cite{23} \cite{18} \cite{14}. Currently, convolutional neural networks (CNNs) are the most effective models for detecting objects \cite{4} \cite{17}. AlexNet's groundbreaking performance in the \textit{2012 ImageNet Large Scale Visual Recognition Challenge} \cite{3} marked a turning point for the image recognition challenge. When applied to the RGB domain, CNNs greatly improved the efficiency of object identification. Girshick et al. \cite{4} introduced labeled Regions with CNN (R-CNN) whose training process is quite time-consuming. Fast R-CNN \cite{5} was presented to deal with issues that could not be solve dby R-CNN. Faster R-CNN \cite{6}, developed by Shaoqing Ren et al., is an object identification system that removed the selective search strategy and allowed the network to learn region proposals. In contrast to region-based algorithms, You Only Look Once (YOLO) is an approach to object detection \cite{7}. It works similarly to DNN-based regression \cite{8}, predicting the existence of an item and its bounding box for a fixed-size grid that tiles the input picture. It requires a single run through the network for detection. 

The use of such frameworks for IR images is hindered by the scarcity of large datasets and the expensive cost of an IR camera. We thoroughly searched the literature but saw only a few number of studies that addressed the drone detection at night utilizing IR images with extreme weather conditions. IR-based UAV datasets captured in outdoor settings with poor atmospheric quality, particularly at night, was also lacking. In this paper, we perform multi-model UAV detection based on IR images captured at night using YOLOv5 and proposed \textbf{Tiny Feature Network (TF-Net)} in complex backgrounds. The main contributions of this paper are listed below:
\begin{itemize}  
\item{1.}To efficiently detect UAVs during the night, we employ UAV detection based on IR images. Relatively less work has been done on IR-based object detection in the literature. To the best of our knowledge, this is the first study considering IR-based UAV detection using proposed TF-Net, YOLOv5s, YOLOv5m, YOLOv5l and YOLOv5n.
\item{2.} In complex backgrounds, it becomes quite challenging to detect any type and size of UAV. We address this problem by making a dataset that consisted IR based UAV images with multiple complex environmental background and weather conditions.
\item{3.} For performance enhancement, we proposed TF-Net, an improved version of YOLOv5s by the introduction kernel based modifications in the neck and backbone. 
\item{4.} For a fair comparison, we trained all considered models with same dataset, learning rate, warmup setup and epochs with early stopping.
\item{5.} The proposed TF-Net detects multi-size and muti-type UAVs with increased precision. Using TF-Net improves the results in terms of precision, mAp@0.5, $IoU$, model size, and, GFLOPs compared to the baseline YOLOv5 models.
\end{itemize}

\section{Literature Review} \label{sec2}
UAVs have grown in popularity for a wide range of uses. Many incidents involving the illegal use of UAVs is making it quite difficult to regulate them and mitigate the associated privacy risks. Real-time drone identification is urgently needed to shield high-security places from the risk of drone intrusion. Currently, there are two major challenges to drone identification in real time. One of them is the rapid speed with which drones move, which calls for equally rapid detection systems. Secondly, multiple sizes of drones further complicate the detection process. A novel approach for detecting multi-rotor drones was presented in \cite{9}. They replaced the Yolov5s backbone with Efficientlite, which allowed them to streamline the model by eliminating unnecessary parameters. To compensate for the accuracy loss, adaptive spatial feature fusion was introduced in the head block to enable the fusion of feature maps with varying spatial resolutions. To speed up network convergence, an angle limitation was added to the existing regression loss function. They trained and validated the model using a dataset of 1,259 multi-rotor UAV images. The modified Yolov5s displayed enhanced detectiion performance in terms of precision, recall, and mAp, at 92.32\%, 89.52\%, and 91.76\% respectively. \cite{10}, evaluated RetinaNet, fully convolutional one-stage object detector (FCOS) , YOLOv3, and YOLOv4. They reduced the size of YOLOv4's convolutional channels and shortcut layer and improved the accuracy of small drone detection. 10,000 drone images were captured with Oneplus phone camera. Results showed that the improved-YOLOv4 model achieved 90.5\% mAP. Authors in \cite{11} taried, CNN, support vector machine (SVM), and k-nearest neighbor (KNN) to train a drone-bird dataset and achieved a maximum accuracy of 93\%.

For detecting low-altitude UAVs, authors in \cite{12} found that YOLOv4 was more effective than YOLOv3 in terms of both accuracy and speed. Due to lack of publicly available low-altitude dataset, they created their own by using three drone models—the DJI-Phantom, Inspire, and XIRO-Xplorer and combined their data with online drone images. They trained the YOLO models with a batch size of 64, 0.9 momentum, 0.0005 decay and 100,000. YOLOv4 achieved 89.32\% accuracy which was 5.18\% better than YOLOv3. To address the security problems posed by UAVs, \cite{13} differentiated two kinds of drones and separated them from birds. They trained their model using a dataset of 10,000 visual images having many drone types like multi rotors, helicopters, and birds. The trained model achieved an accuracy and mAp of 83\% and 84\%, respectively. Schumann et al. \cite{14} used median-background subtraction in conjunction with a deep learning-based region proposal network (RPN) and  VGG-conv5 as a classifier. Their dataset had 10,286 images and won first place in the 2017 drone vs. bird competition. In \cite{15}, background-subtraction method was used for the identification of the moving items. They classified the detected items into three categories: bird, drone, and background using MobileNet-v2. This method produced encouraging results with accuracy, recall, and f-score of 70.1\%, 78.8\%, and 74.2\% respectively. 

An anti-drone system with automatic identification for the attacking drones was presented by Khan et al. \cite{16}. Radar which emitted microwaves is used to find moving objects and then YOLOv3 was used to identify the category of object identified by the activated the camera. If the identification confidence was more than 75\%, then the laser gun was used to lock the identified object. Study in \cite{17} fine-tuned YOLOv5 model on a publicly accessible dataset of 1359 images. They dataset was also augmented to overcome the missing data points. They compared to results with baseline standard YOLOv3, YOLOv4, and maskRCNN. The suggested technique was more effective with 95.2\% accuracy. The work in \cite{18} used Darknet as the YOLOv4 backbone for UAV and bird detection. The proposed approach successfully overcame the major difficulty of identifying comparable tiny objects near and distant in all situations, with 98.3\% detection accuracy. 

Images captured by thermal cameras are unaffected by smoke or other atmospheric factors, making them an indispensable tool for use in search-and-rescue operations and fire prevention. In \cite{19}, a Faster RCNN was used to examine thermal and visual spectrum images side by side. For object recognition and classification at the sea's surface, authors in \cite{20} studied UAV thermal pictures for easy identification and location of aquatic objects. The experimental findings revealed a 92.5\% accuracy across a testing dataset. The study in \cite{21} presented a multi-sensor fully autonomous drone detection system with a thermal infrared camera. The proposed system reduced the false positives by employing sensor fusion to make it more reliable than individual systems. To tackle the unavailability of the datasets in this domain, this paper also presented annotated video datasets with 650 visible and IR videos of birds, airplanes, helicopters, and drones.

\section{Proposed TinyFeatureNet (TF-Net)}
The proposed TF-Net has four blocks: input, backbone, neck, and head. To facilitate speedy inference with no mAP cost, the \textit{input block} maps spatial information from the input imagery to the channel dimension. It also performs data preparation by using mosaic data augmentation (MDA) and adaptive image filling (AIF) techniques. Training time and the need for a tiny mini-batch size is reduced due to MDA's ability to teach the TF-Net object identification on a lesser scale than conventional object detection methods. The cross-stage partial network (CSP) and spatial pyramid pooling (SPP) are used by the \textit{backbone block} to extract feature maps of varying sizes from the input pictures. The calculation and inference times were further reduced by using the BottleneckCSP block. For precise detection, SPP helped by the extraction of feature maps at three different scales. Feature pyramid network (FPN) is used in \textit{neck block}, which captures semantic attributes from the highest to lowest levels of the hierarchy. The final detection is provided by the last  \textit {head block}.
Anchor boxes are sets of bounding boxes that have been labeled in terms of their height and width. In order to record the scale and aspect ratio of different object classes, these boxes are selected depending on the item sizes in training datasets. In TF-Net, we used 8 $\times$8 (P3), 16 $\times$ 16 (P4) and 32 $\times$ 32 (P5) anchor boxes for the identification of target objects.
\begin{figure}[h]
  \begin{center}
  \includegraphics[width=5 in]{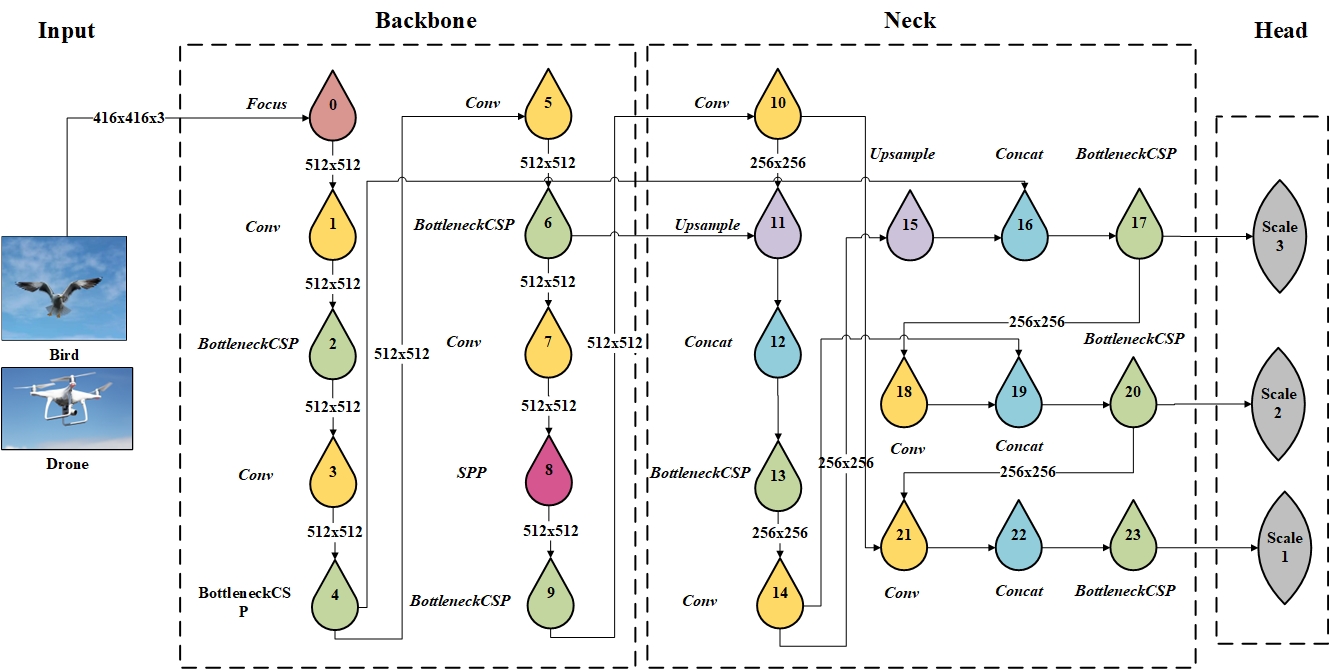}
  \caption{Proposed \textbf{TF-Net Model}.}
  \label{fblock}
  \end{center}   
\end{figure}
Overall TF-Net uses a single neural network to analyze the whole image. It partitions the input image of 416$\times$416$\times$3 into grids and provides probabilistic bounding box predictions for each grid cell. The projected probability then weights the bounding boxes with precise predictions after only one forward propagation through the neural network. Then the max suppression strategy guarantees that the TF-Net only performs one identification for each item. The size of the kernels used for the head and backbone impacts the extracted features, training parameters, gradients, and GFLOPs. The detailed TF-Net model network structure is shown in Fig \ref{fblock}. Proposed TF-Net determined the best kernel size for feature maps which excellently performed UAV detection during night time in bad weather conditions. The model depth multiple and layer channel multiples are 0.33 and 0.50, respectively. 

TF-Net uses cross-entropy function for a cost function. To prevent the model from over-fitting, we added mixed regularization $L1$ and $L2$ to the loss function. Adding mixed regularization makes the model more robust for night time detection. Loss function $L$ is optimized as follows:
\begin{equation}
\begin{gathered}
\text { C\_L }=\mathrm{G}(\mathrm{a}, \mathrm{b})=-\sum_{\mathrm{i}} \mathrm{a}(\mathrm{i}) \ln \mathrm{b}(\mathrm{i}) \\
\mathrm{L} \text { = C\_L }+\sum_{\mathrm{j}}\left(\alpha\left|\psi_{\mathrm{j}}\right|+(1-\alpha) \psi_{\mathrm{j}}^2\right),
\end{gathered}
\end{equation}
where $a$ and $b$ represent the actual and predicted probability distribution of $x$, respectively, $\alpha$ represents the regularization parameter and $\psi$ represents weights. Before computing $L1$ and $L2$ mixed regularization, the average cross-entropy of the whole batch is computed to avoid over-fitting.

\section{Dataset and Implementation}
\begin{figure}[h]
\centering
\setkeys{Gin}{width=0.4\linewidth}
\includegraphics{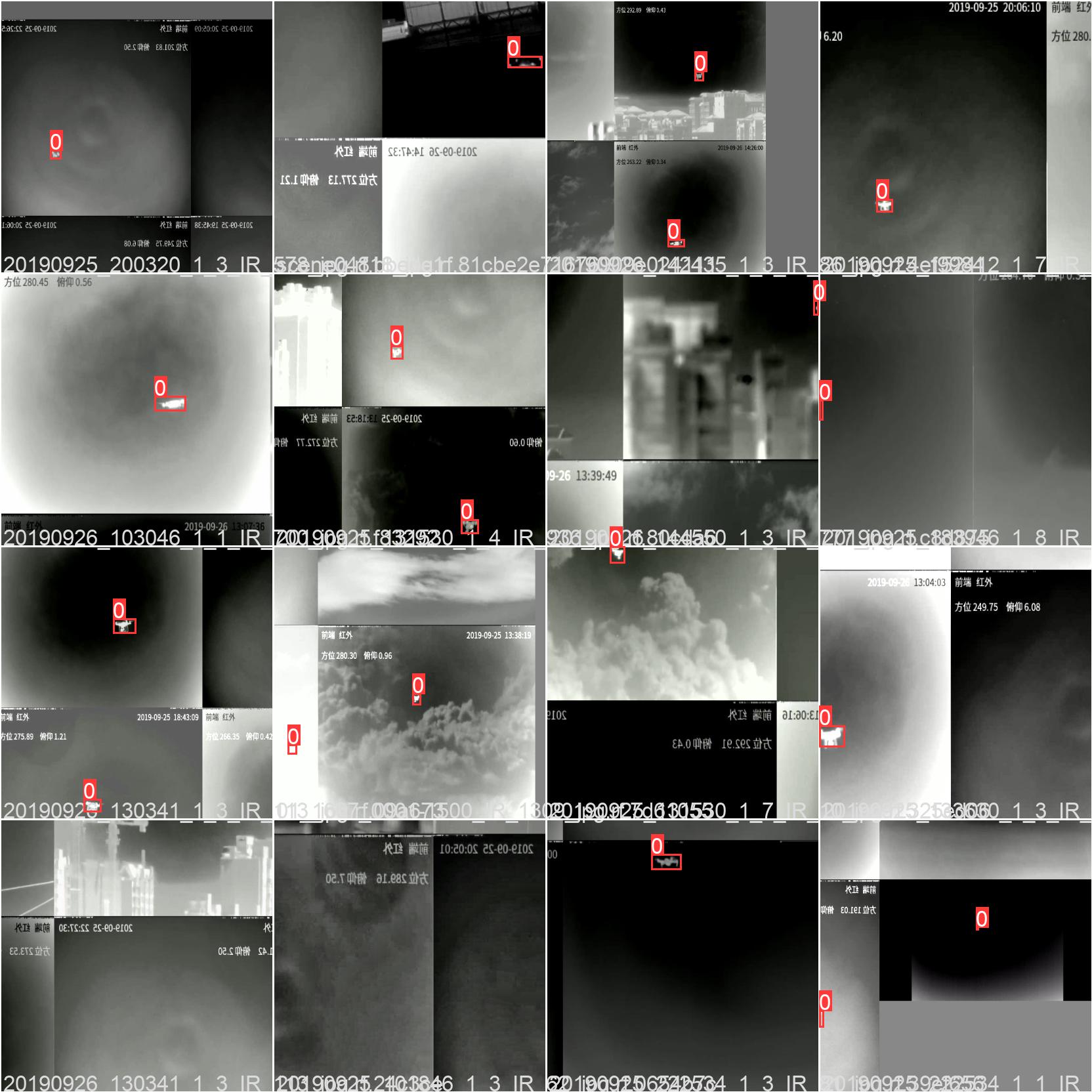}
\caption{Data Set (a) Distribution (b) Ground-truth images.} 
\label{f4}
\end{figure}

The detection of drones is quite difficult at night because it gets harder to see them with nake eye. In this study, we assessed the performance of TF-Net and YOLOv5s,m,l and n for nighttime UAV detection. We consider a dataset with more than 3000 images of UAVs flying at night in various backgrounds \cite{r2}. We set the train-test ratio to 8:2 with 3100 training  and 791 test images. Dataset visualization  shown in Fig. \ref{f4}a depicts the multi-size UAVs target. YOLO based object detection models need datasets with class categories, bounding boxes, and annotation files. We used $Roboflow$ \cite{r1}, an open-source dataset platform, to create a dataset that had all the necessary files for the TF-Net and YOLOv5 models. Initially, the images are pre-processed, by resizing them all to a fixed dimension size of 416$\times$416 scaling, contrast enhancement before training.
\begin{table}[h]
\caption{Hyper-parameters qualitative analysis}
\centering
    \begin{tabular}{l l l l l l}
    \hline
    \textbf{Input size} &\textbf{Epochs}  &\textbf{Layers} & \textbf{Learning rate} &\textbf{Momentum} & \textbf{Weight Decay}
    \\ \hline
    416$\times$ 416 &300  & 232 &0.01 &0.937 &0.0005
    \\ \hline
  \label{t3}
    \end{tabular}
\end{table}
We configure the hyper-parameters as described in Table. \ref{t3} to have smooth data training without higher loss and over-fitting. Adam's and the SGD optimizer's initial learning rate (lr0) was set to 0.1. The one-cycle learning rate (lrf) was changed to be 0.1. The momentum for the SGD optimizer was set at 0.937 with a weight decay of 0.0005. At the warmup epoch of 3.0, the initial warmup momentum is 0.8 and the initial warmup bias is 0.1. Box loss gain, class loss gain, object loss gain, and focused loss gamma all have initial values of zero. The multiple threshold is 4.0, and the $IoU$ training threshold is 0.20. All models are trained on Google Colab with a K80 GPU and 12GB RAM. During training, the objectness loss decreased from 0.019 to 0.014 and the box loss decreased from 0.09 to 0.04. The classifier head is trained to identify the type of the target item using a classification loss. Values range from 0.035 to 0.010. An enhanced classification loss has been seen for the suggested approach. The precision, recall, mAp, and $IoU$ over 100 plus epochs show an increasing trend, which implies that the model's learning patterns are going well. These results still have room for improvement, which will be addressed in the future work.
\section{Results and Discussions }
YOLOv5l has a depth multiple (DM) of 1.0 and  a width multiple (WM) of 1.0. YOLOv5m shows 0.67 DM and 0.75 WM, YOLOv5s and TF-Net has 0.33 DM with a WM of 0.50 while YOLOv5n has 0.33 DM with a WM of 0.25. YOLOv5l is the deepest model and takes the largest to train due to its increased convolution and computational complexity while YOLOv5n takes the least and has the lowest computational complexity but the evaluation metrics computed for both these are lower as compared to YOLOv5s and TF-Net. We assess the performance of the proposed TF-Net and YOLOv5 (s,ml and n) by calculating true positive (TP), true negative (TN), false positive (FP), and false negative (FN), precision (P), recall (R), mean accuracy precision (mAp), and intersection over union ($IoU$).

\subsection{ Performance evaluation }
\begin{figure}[h]
     \centering
     \begin{subfigure}[b]{0.45\textwidth}
         \centering
         \includegraphics[width=\textwidth]{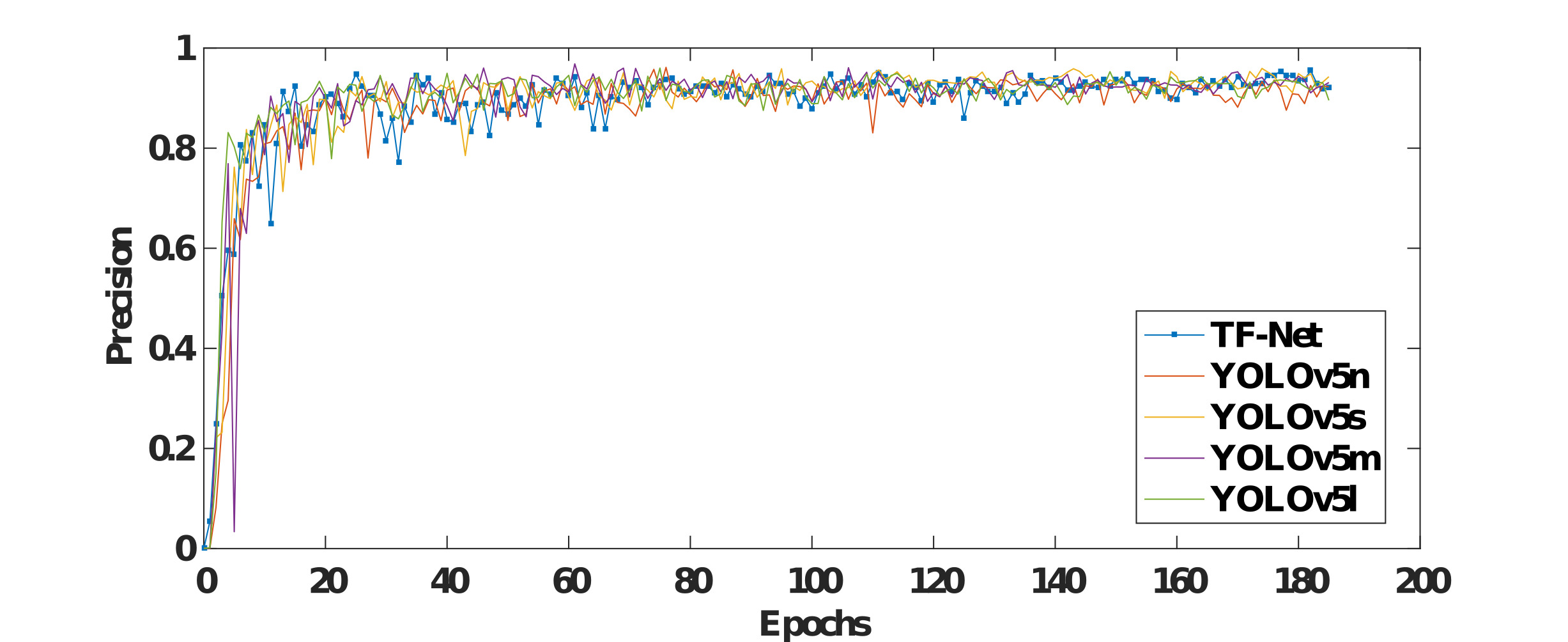}
         \caption{Precision}
         \label{fig:a}
     \end{subfigure}
     \begin{subfigure}[b]{0.45\textwidth}
         \centering
         \includegraphics[width=\textwidth]{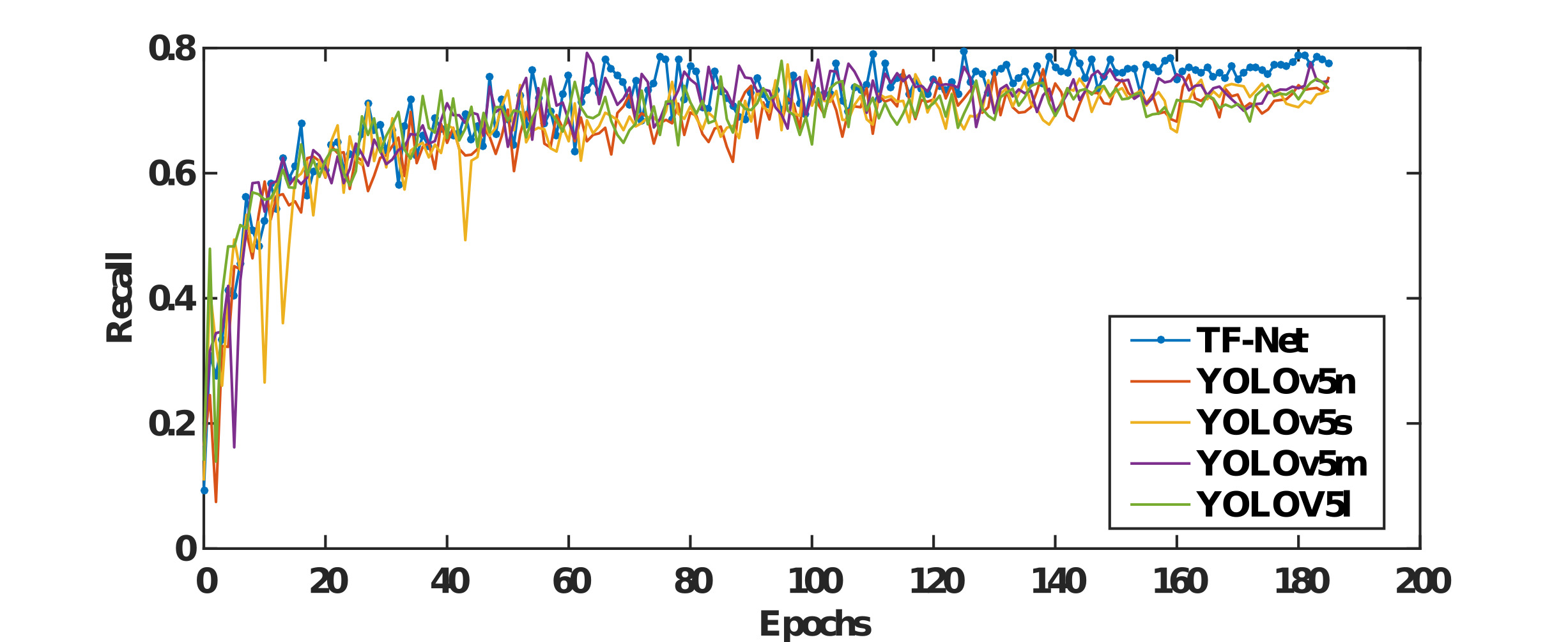}
  \caption{Recall}
         \label{fig:b}
     \end{subfigure}
          \begin{subfigure}[b]{0.45\textwidth}
         \centering
         \includegraphics[width=\textwidth]{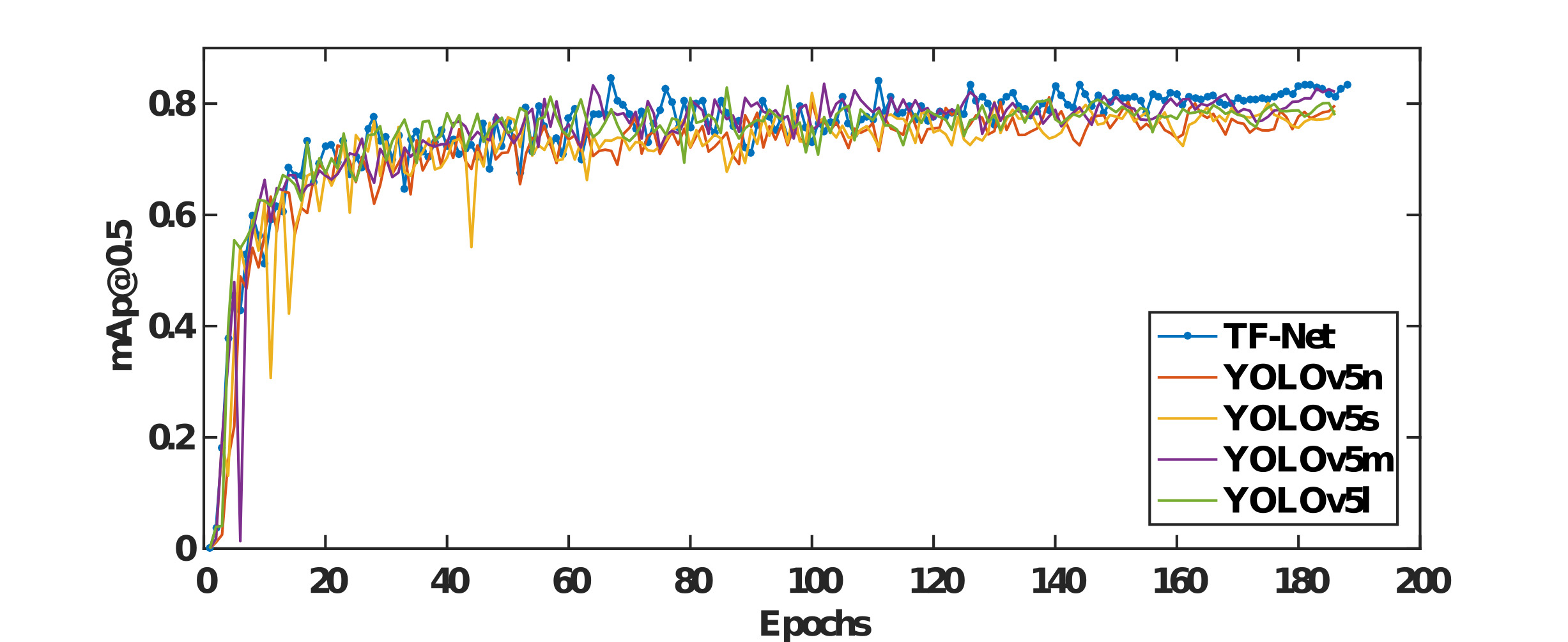}
         \caption{mAP}
         \label{fig:c}
     \end{subfigure}
          \begin{subfigure}[b]{0.45\textwidth}
         \centering
         \includegraphics[width=\textwidth]{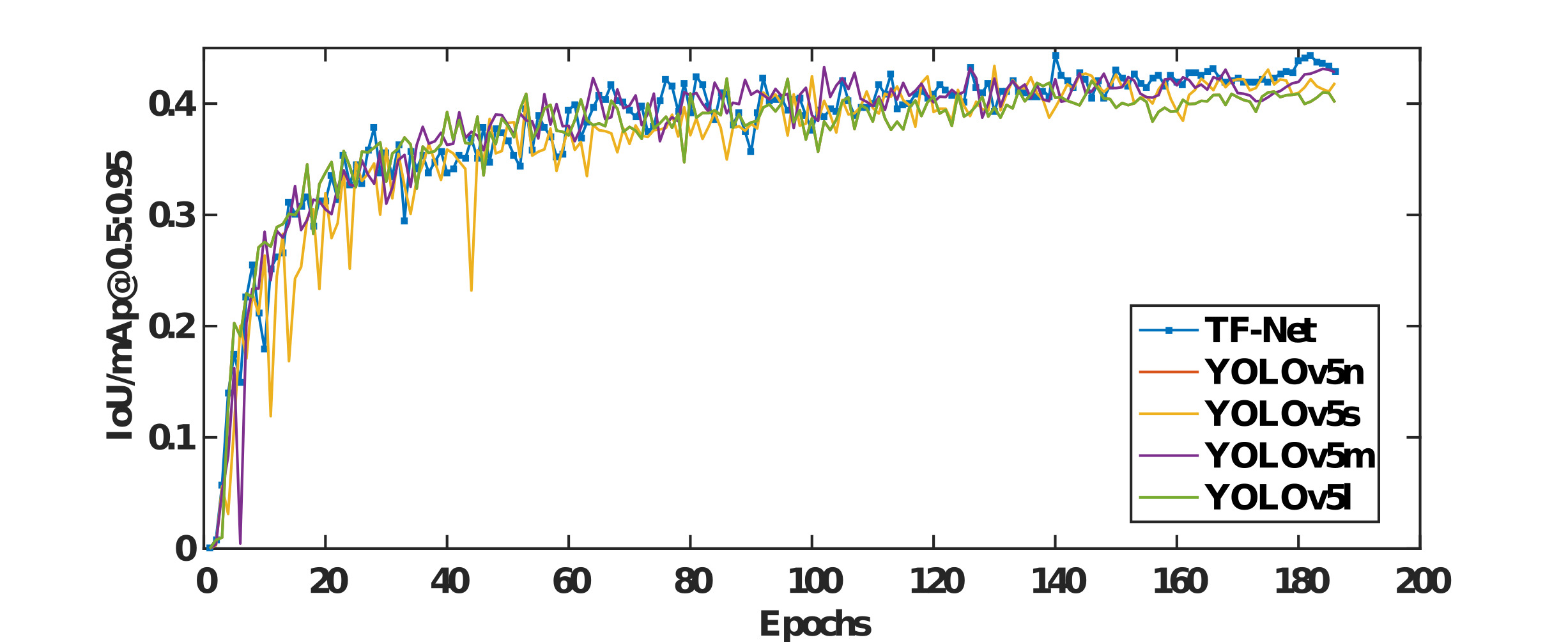}
         \caption{IOU}
         \label{fig:d}
     \end{subfigure}
     \caption{Graphical representation of the merged dataset.}
        \label{f3}
\end{figure}

YOLOV5n, YOLOv5s, YOLOv5m, and YOLOv5l achieved a precision of 92.3\%, 93.1\%, 91.2\%, and 90.8\%, respectively. YOLOv5l had the lowest precision among all the models which shows its decreased sensitivity for UAV detection in night. In terms of recall, YOLOv5m performed well by achieving 78.1\% compared to the rest of the YOLOv5 models and TF-Net. The highest precision of 93.1\% is achieved by YOLOv5s. Similarly, YOLOv5m has the highest mAp@0.5 of 83.3\% while YOLOv5n has the highest $IoU$ of 43.5\%.  However, TF-Net achieved the best of all when it comes to precision, mAp@0.5, and $IoU$ of 95.6\%, 84\% and 44.8\% respectively. This makes TF-Net best suited for multiple-size UAV detection in complex backgrounds during night. Fig. \ref{fig:a}  \& \ref{fig:b} displays the recall and precision results over 180 epochs for both TF-Net and YOLOv5. These graphs show an increasing trend, hence proving the model's correctness.

The proposed TF-Net outperforms all the YOLOv5 models (s,m,l,n) in terms of precision, mAp@0.5, and $IoU$ as shown in Table \ref{T1}. TF-Net achieved 95.7\% precision which is 3.4\% higher than YOLOv5n, 2.1\% higher than YOLOv5s, 4.5\% higher than YOLOv5m and 4.9\% higher than YOLOv5l.YOLOv5s achieved 1\% higher recall than TF-Net. Compared to other YOLOv5 models, TF-Net achieved 77\% recall which is 1.4\% higher than YOLOv5n, 0.7\% higher than YOLOv5m and 2.1\% higher than YOLOv5l. TF-Net's, mAp@0.5, and $IoU$ are also higher by 84\% and 44.8\%, respectively. This shows the higher sensitivity of TF-Net for detecting multi-size objects compared to the rest of the standard YOLOv5 models during night time. 
\begin{table}[h]
\centering
\caption{Training Evaluation Metrics}
\begin{tabular}{l l l l l l l}
\hline
        \textbf{Model} & \textbf{TP} & \textbf{FN} &\textbf{Precision  (\%)} & \textbf{Recall  (\%)} & \textbf{mAp @0.5 (\%)} & \textbf{IoU (\%)}   \\ \hline
        \textbf{TF-Net } & 82& 18& \textbf{95.7} & 77.4 & \textbf{84} & \textbf{44.8 }  \\ \hline
      
        YOLOv5n & 80	($\downarrow$ 2)& 20 ($\uparrow$ 2)&92.3 ($\downarrow$ 3.4) & 76 ($\downarrow$ 1.4) & 80.5 ($\downarrow$ 3.5) & 43.5 ($\downarrow$ 1.3 )  \\ \hline
        
        YOLOv5s & 81 ($\downarrow$ 1)	&19 ($\uparrow$ 1)& 93.1 ($\downarrow$ 2.6) & \textbf{78.4 }($\uparrow$ 1 ) & 80.3 ($\downarrow$ 3.7) & 43.4 ($\downarrow$ 1.4)  \\ \hline
        
        YOLOv5m  & \textbf{83} ($\uparrow$ 1)	&\textbf{17 }($\downarrow$ 1) & 91.2 ($\downarrow$ 4.5)& 78.1($\uparrow$ 0.7) & 83.6 ($\downarrow$ 0.4) & 43.3 ($\downarrow$ 1.5) \\ \hline
        
        YOLOv5l& 77 ($\downarrow$ 5)	&23 ($\uparrow$ 5) & 90.8 ($\downarrow$ 4.9)& 75.3 ($\downarrow$ 2.1)& 82.9 ($\downarrow$ 1.1) & 42.2 ($\downarrow$ 2.6) \\ \hline
        
    \end{tabular}
    \label{T1}
\end{table}

After training over 200 epochs, TF-Net has the highest true positive (TP) rate of 82  wich is 2\% higher than YOLOv5n, 1\% higher than YOLOv5s and 5\% higher than YOLOv5l. YOLOv5s yielded slightly less TP rate of 81\%. YOLOv5s has the same number of network layers as TF-Net but the kernel and feature-map sizes are different. TP rate is also referred to as $"$Sensitivity$"$. This shows that the TF-Net model is most sensitive to multi-size UAV detection and will have excellent sensitivity upon recalling and correctly identifying drones as drones.
\begin{figure}[h]
     \centering
     \begin{subfigure}[b]{0.3\textwidth}
\includegraphics[width=\textwidth]{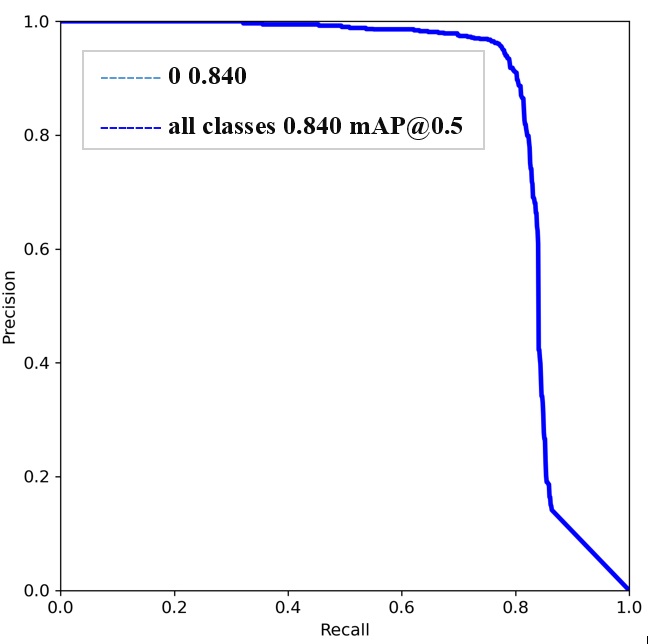}
         \caption{TF-Net}
         \label{f7a}
     \end{subfigure}
     \begin{subfigure}[b]{0.3\textwidth}
         \includegraphics[width=\textwidth]{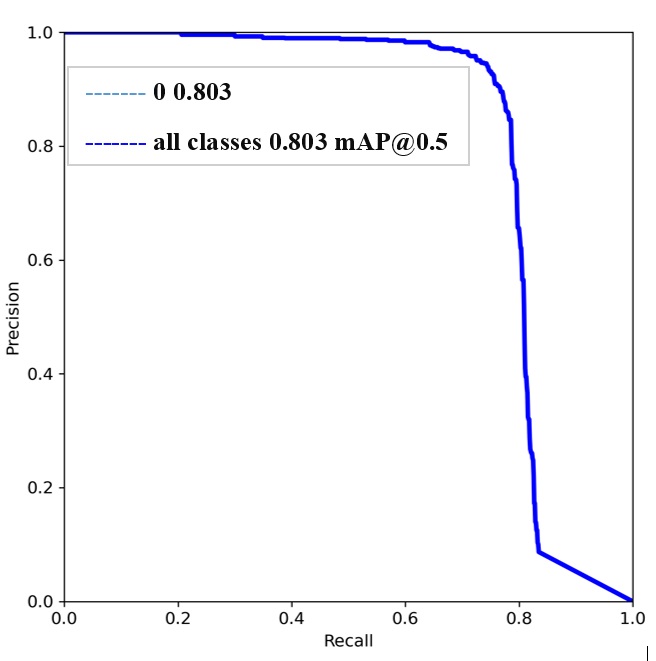}
         \caption{YOLOv5s}
         \label{f7b}
     \end{subfigure}
     \begin{subfigure}[b]{0.3\textwidth}
         \centering
         \includegraphics[width=\textwidth]{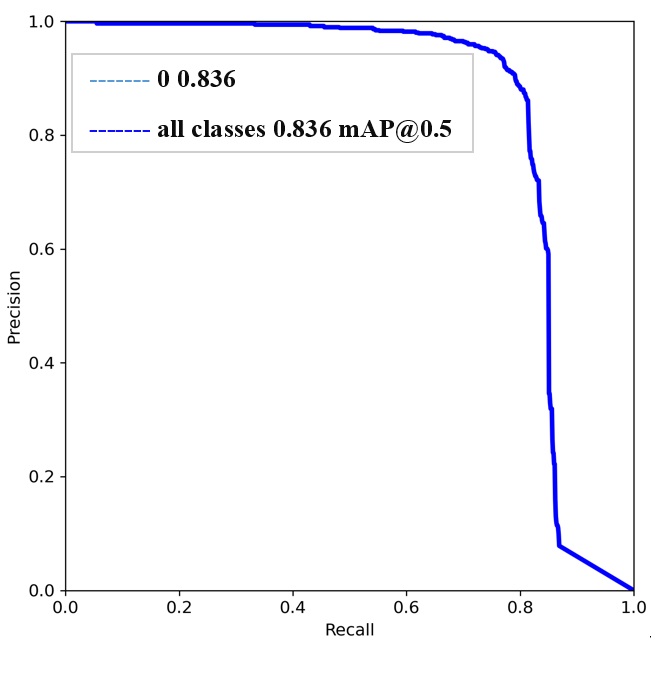}
         \caption{YOLOv5m}
         \label{f7c}
     \end{subfigure}
        \begin{subfigure}[b]{0.3\textwidth}
         \includegraphics[width=\textwidth]{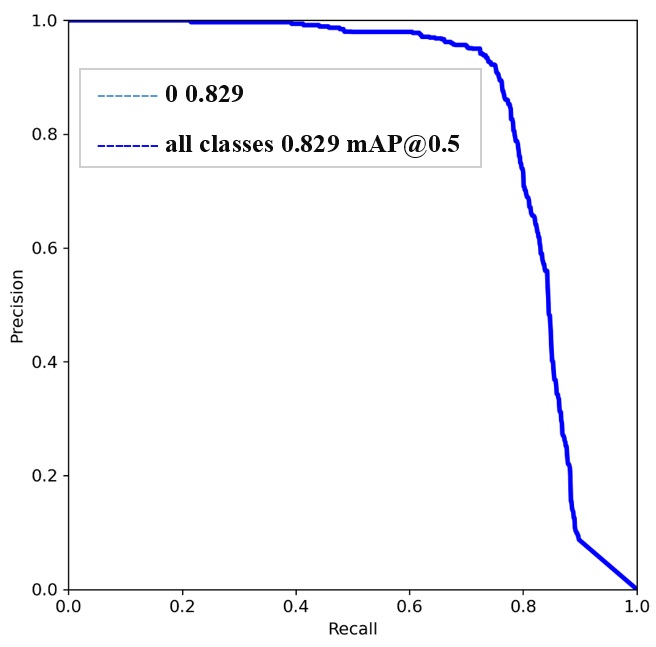}
         \caption{YOLOv5l}
         \label{f7d}
     \end{subfigure}
        \begin{subfigure}[b]{0.3\textwidth}
         \includegraphics[width=\textwidth]{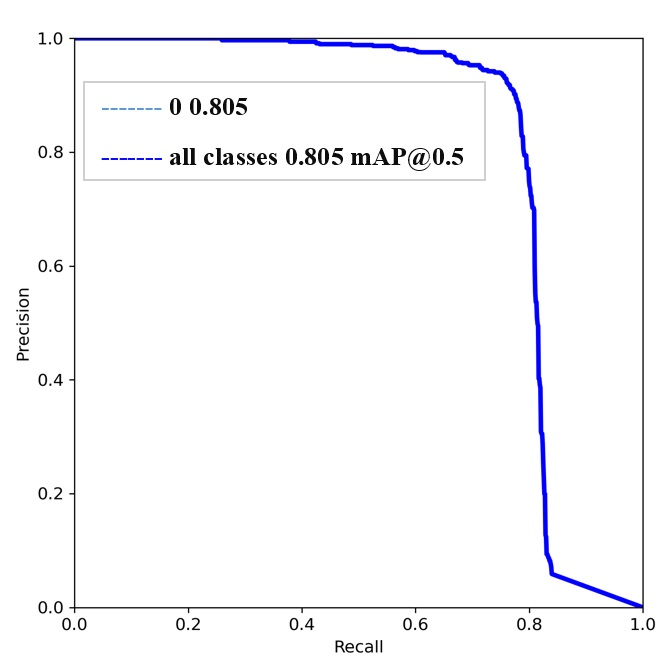}
         \caption{YOLOv5n}
         \label{f7e}
     \end{subfigure}
        \caption{Precision-Recall Curve}
        \label{f7}
\end{figure}

The plot in Fig. \ref{fig:c} \& \ref{fig:d} shows mAp@0.5 and $IoU$ of all trained models, respectively. mAp@0.5 and $IoU$ of the TF-Net have significantly improved compared to the YOLOv5 models. The mAP value compares the ground-truth bounding box to the detected box and returns a score. The higher mAP score of TFNet shows its ability for accurate detection. TF-Net mAp increased by 3.5\% w.r.t YOLOv5n, 3.7\% w.r.t YOLOv5s, 0.4\% w.r.t YOLOv5m and 1.1\% w.r.t YOLOv5. The $IoU$ of TF-Net increased by 1.3\% w.r.t YOLOv5n, 1.4\% w.r.t YOLOv5s, 1.5\% w.r.t YOLOv5m and 2.6\% w.r.t YOLOv5. Although there is always a trade-off between mAp and $IoU$ but in our case, both metrics have shown improvement which specifies that TF-Net excellent ability to overlap between the predicted and ground truth bounding box.
The precision-recall curves shown in Fig. \ref{f7} give the value of mAp@0.5. The precision-recall curve in Fig. \ref{f7a} achieved the highest mAp@0.5 of 84\% for TF-Net. Comparatively, lower mAp@0.5 is achieved by YOLOv5s, YOLOv5m, YOLOv5l, and YOLOv5n.
\begin{figure}[h]
     \centering
     \begin{subfigure}[b]{0.09\textwidth}
         \centering
         \includegraphics[width=\textwidth]{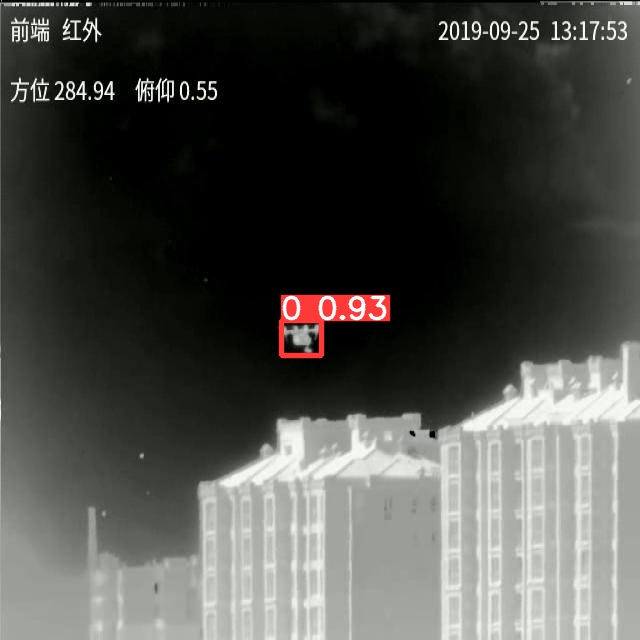}
         \includegraphics[width=\textwidth]{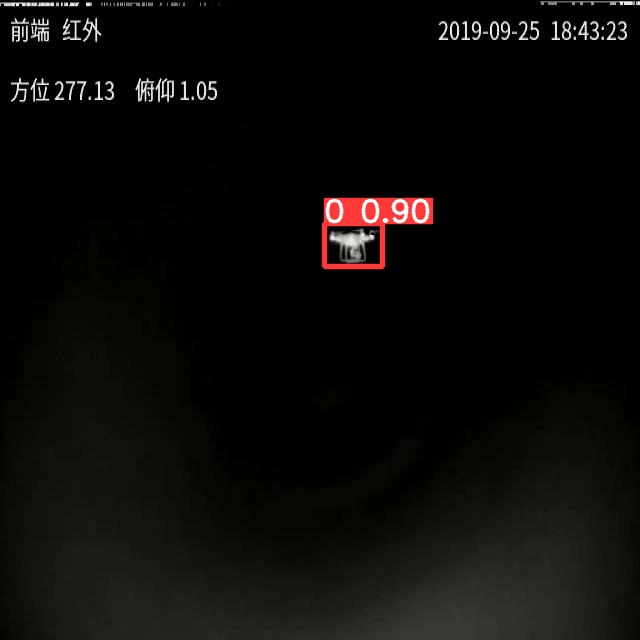}
         \includegraphics[width=\textwidth]{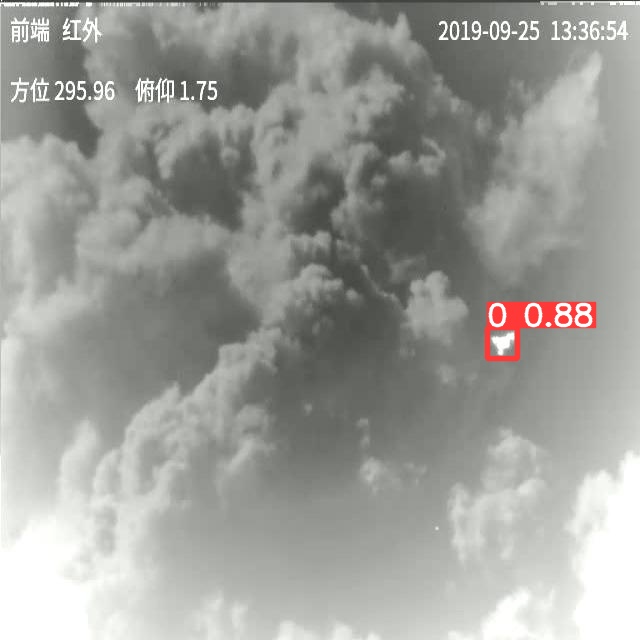}
         \includegraphics[width=\textwidth]{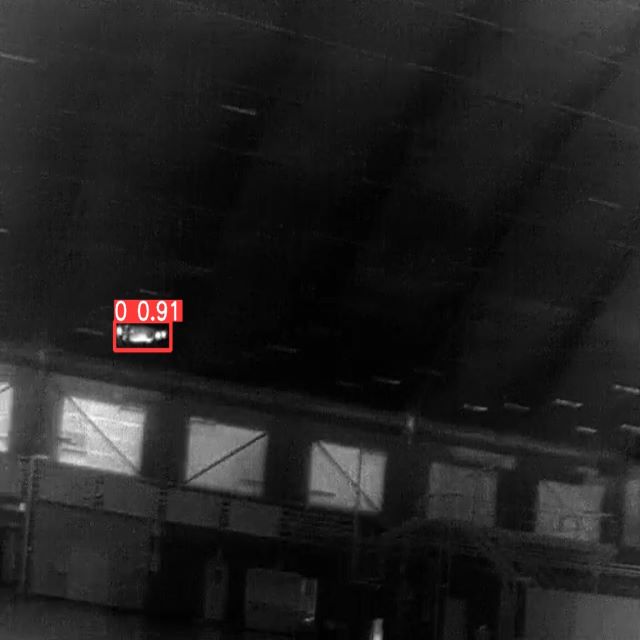}
         \includegraphics[width=\textwidth]{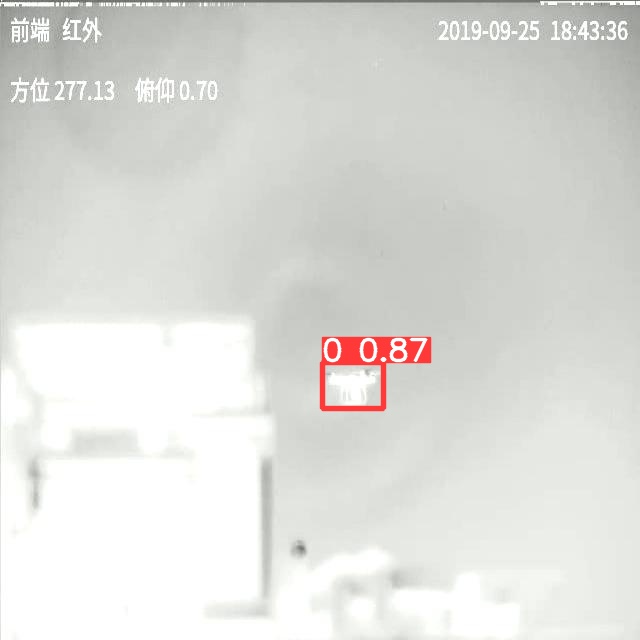}
         \includegraphics[width=\textwidth]{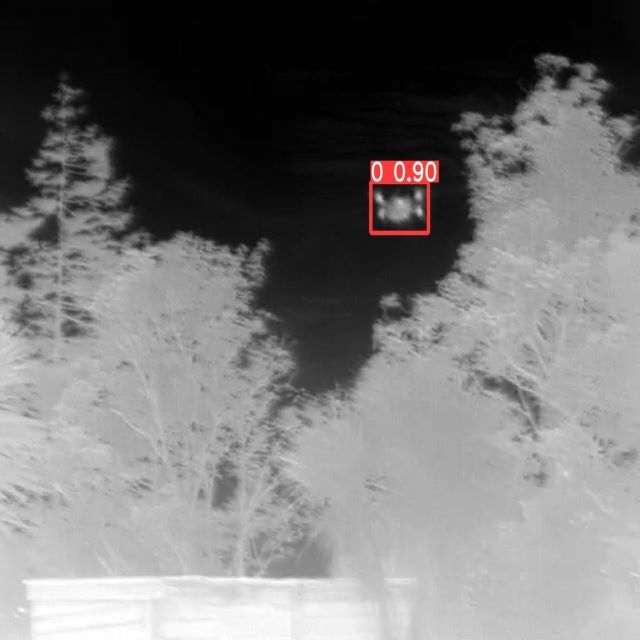}
         \includegraphics[width=\textwidth]{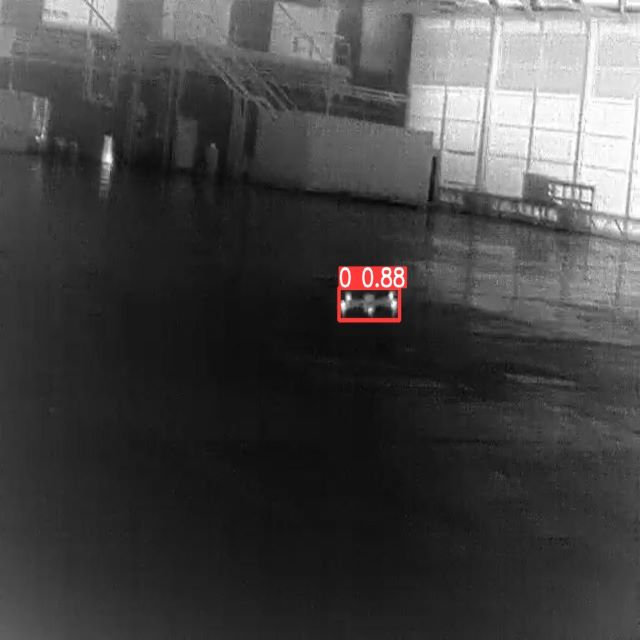}
         \caption{TF-Net}
         \label{figa}
     \end{subfigure}
     \hfill
     \begin{subfigure}[b]{0.09\textwidth}
         \centering
         \includegraphics[width=\textwidth]{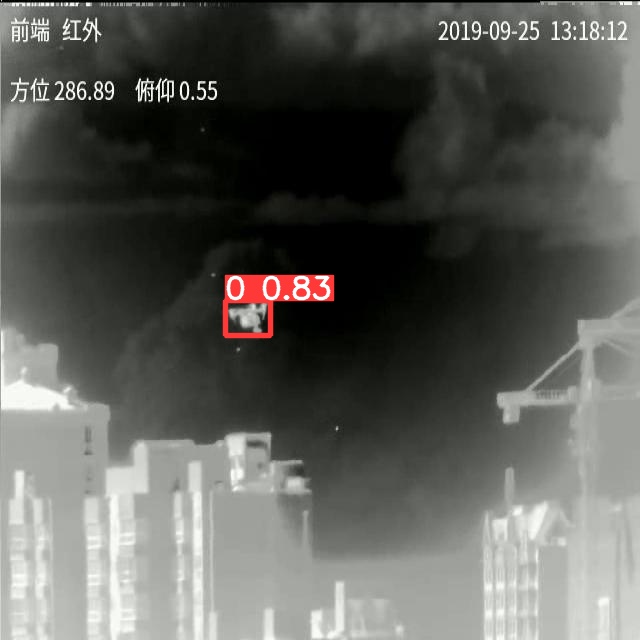}
         \includegraphics[width=\textwidth]{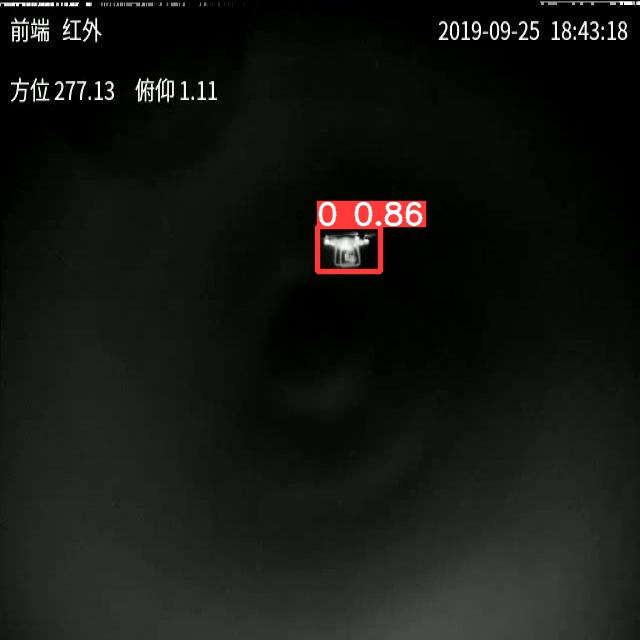}
         \includegraphics[width=\textwidth]{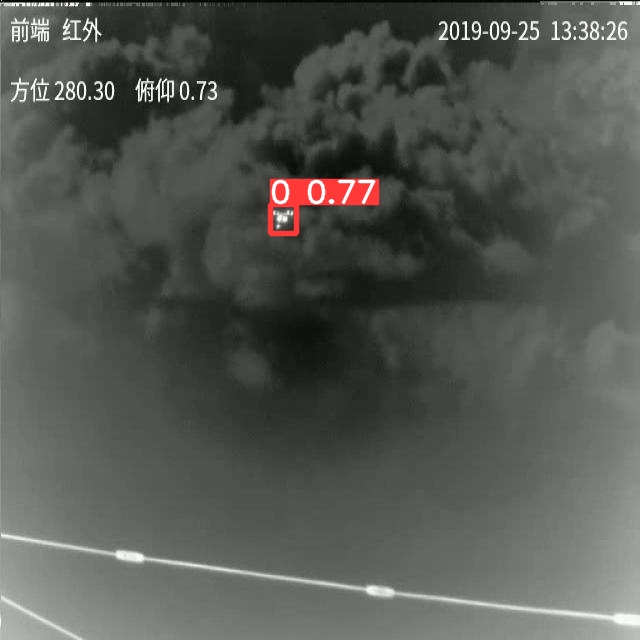}
         \includegraphics[width=\textwidth]{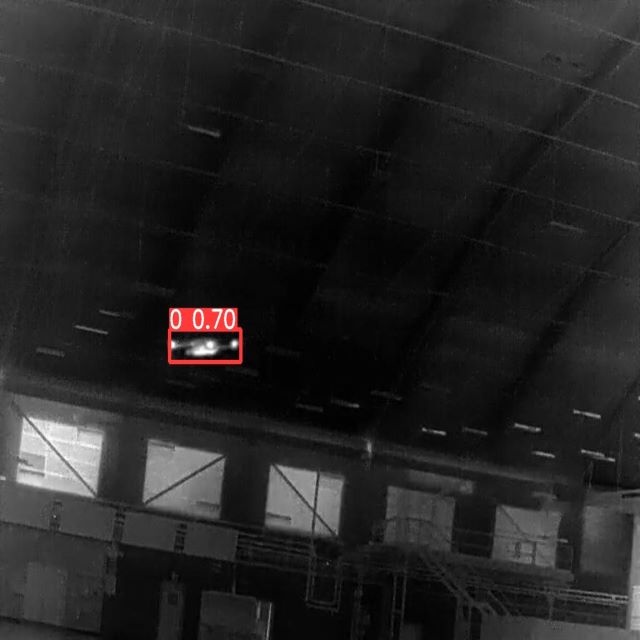}
         \includegraphics[width=\textwidth]{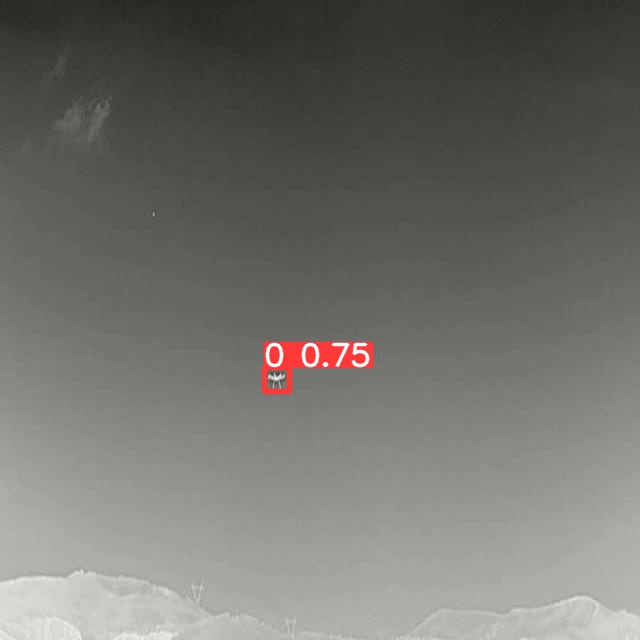}
         \includegraphics[width=\textwidth]{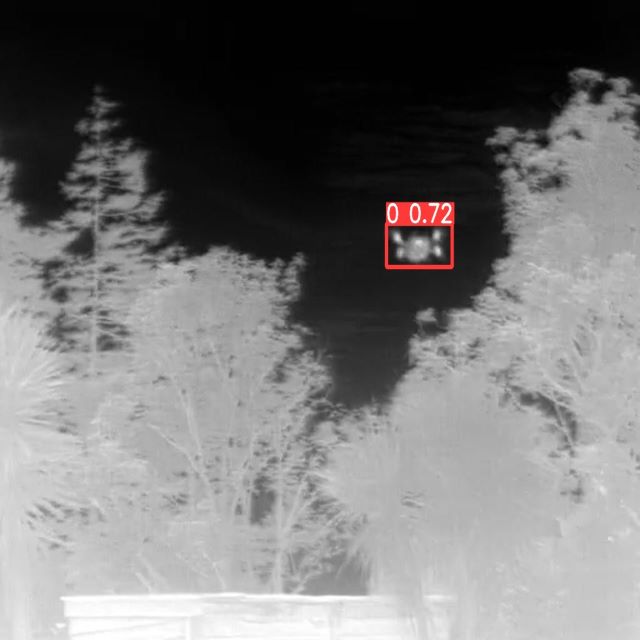}
         \includegraphics[width=\textwidth]{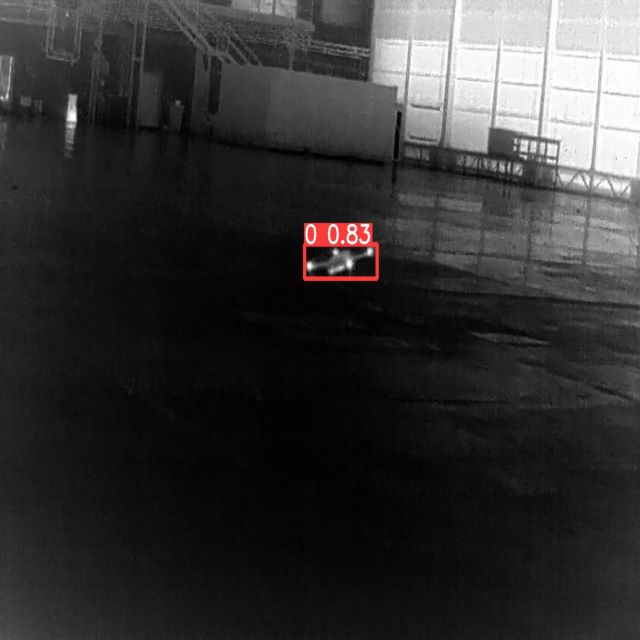}
         \caption{YOLOv5n}
         \label{figb}
     \end{subfigure}
     \hfill
     \begin{subfigure}[b]{0.09\textwidth}
         \centering
         \includegraphics[width=\textwidth]{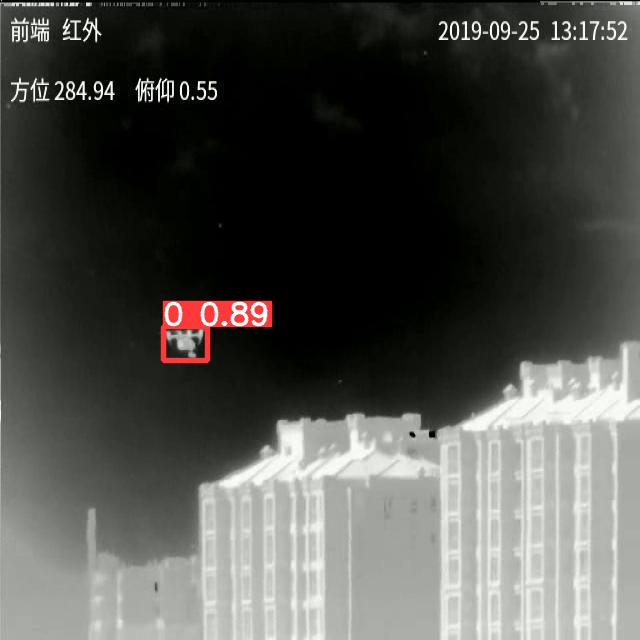}
         \includegraphics[width=\textwidth]{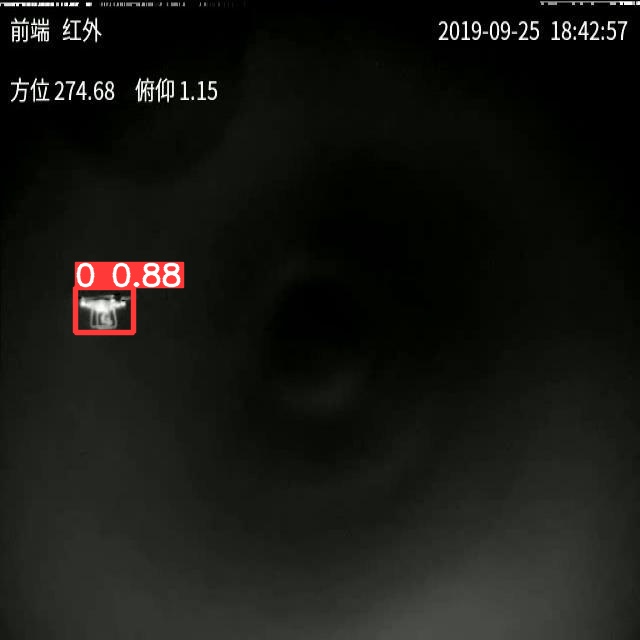}
         \includegraphics[width=\textwidth]{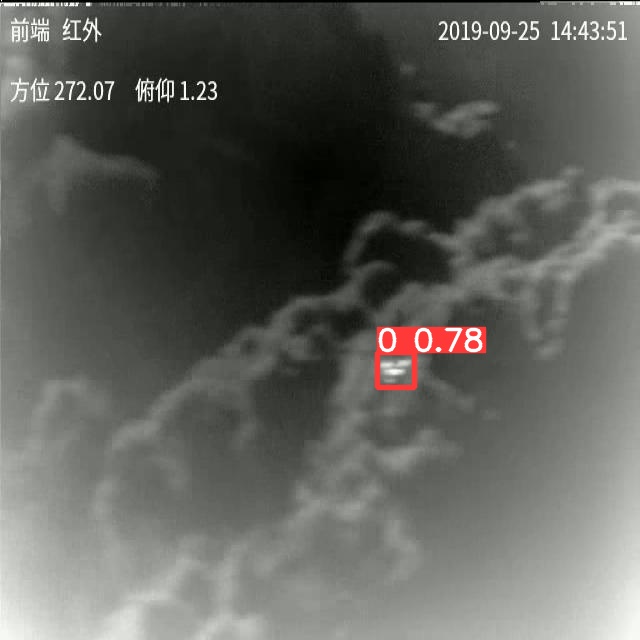}
         \includegraphics[width=\textwidth]{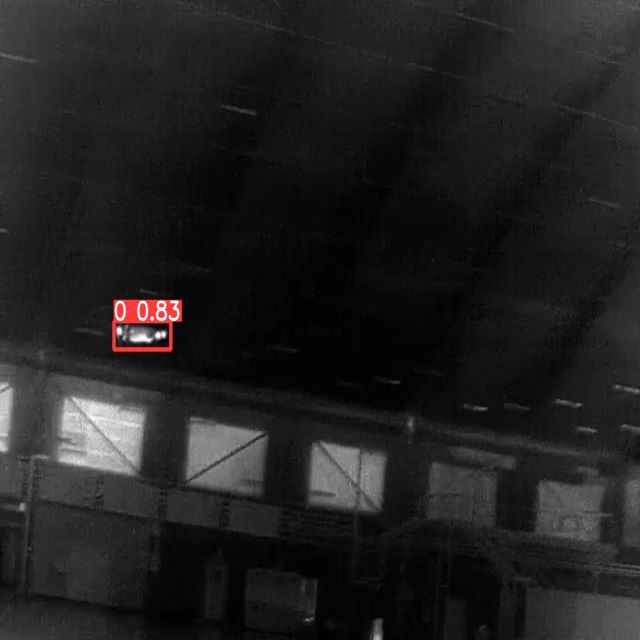}
         \includegraphics[width=\textwidth]{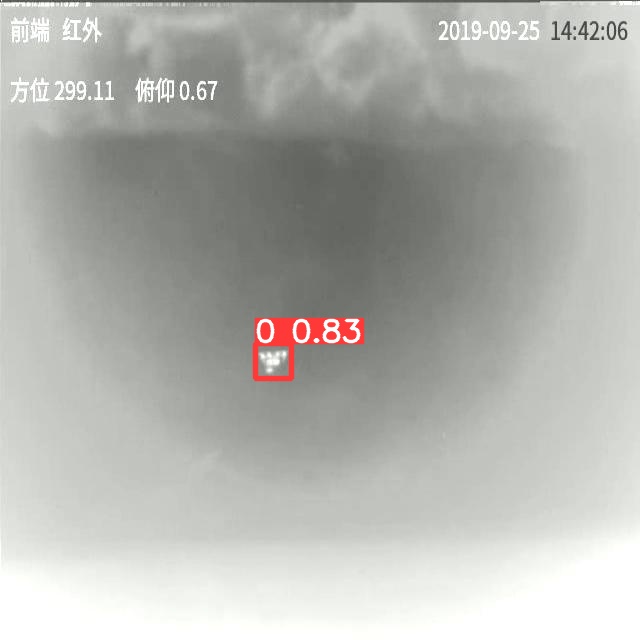}
         \includegraphics[width=\textwidth]{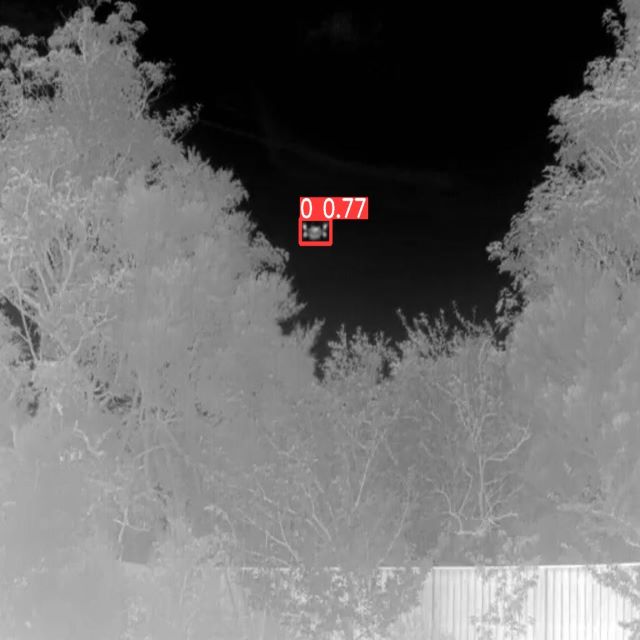}
         \includegraphics[width=\textwidth]{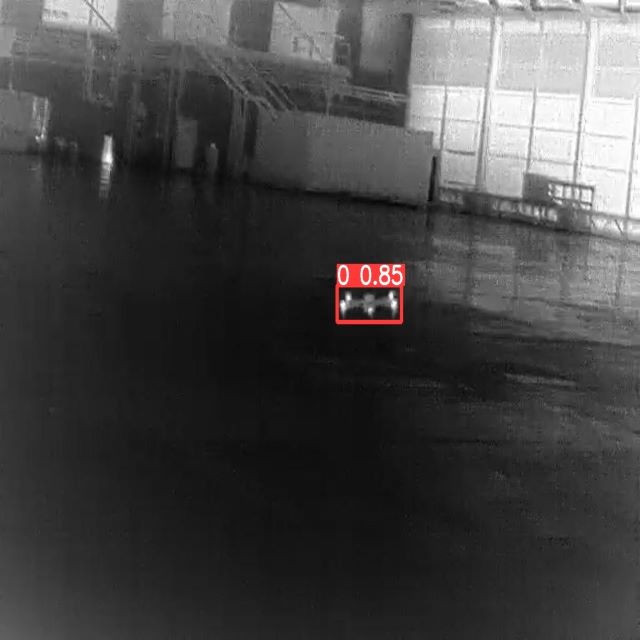}
         \caption{YOLOv5s}
         \label{figc}
     \end{subfigure}
     \hfill
     \begin{subfigure}[b]{0.09\textwidth}     
         \centering
         \includegraphics[width=\textwidth]{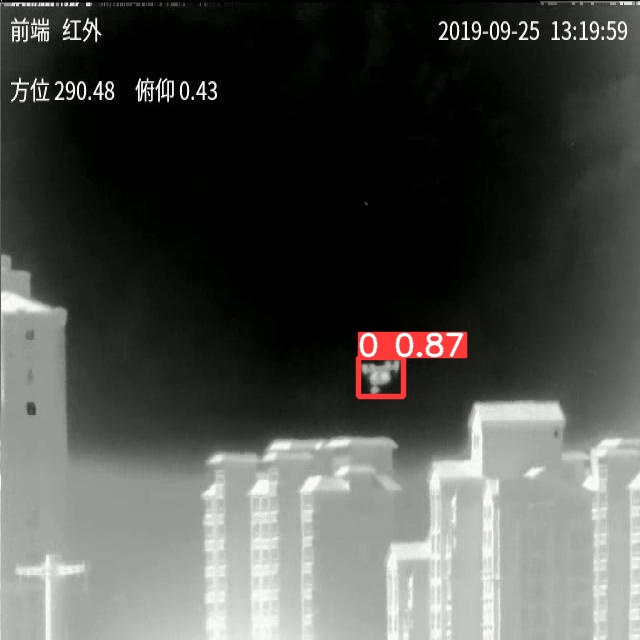}
         \includegraphics[width=\textwidth]{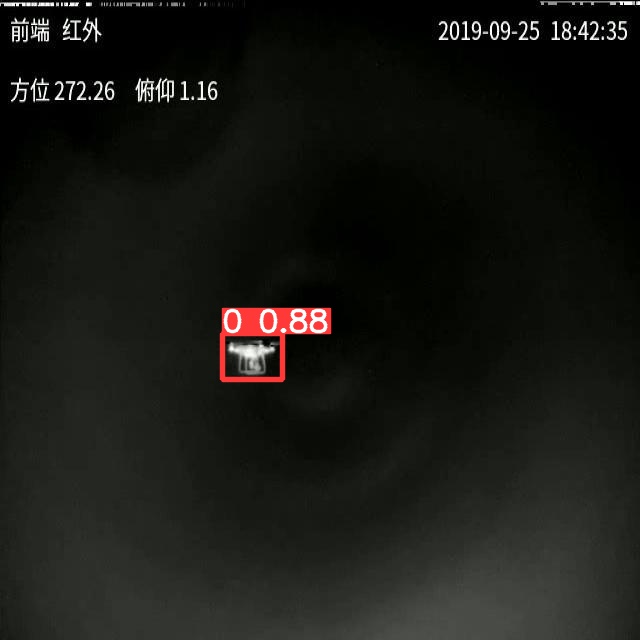}
         \includegraphics[width=\textwidth]{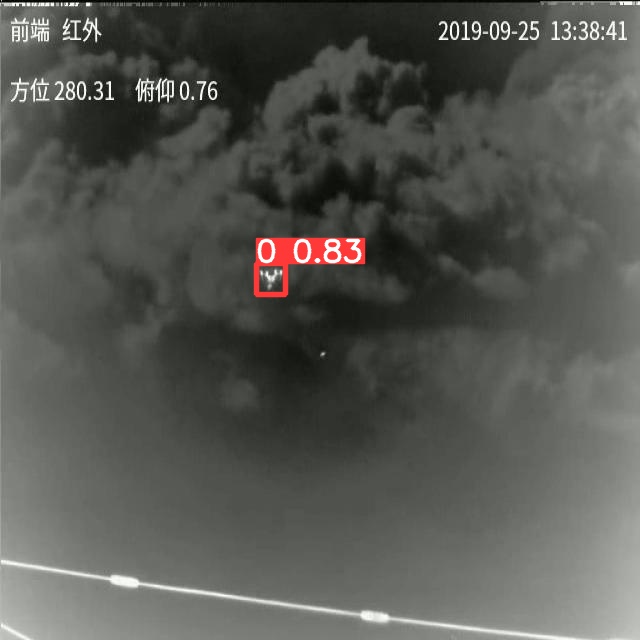}
         \includegraphics[width=\textwidth]{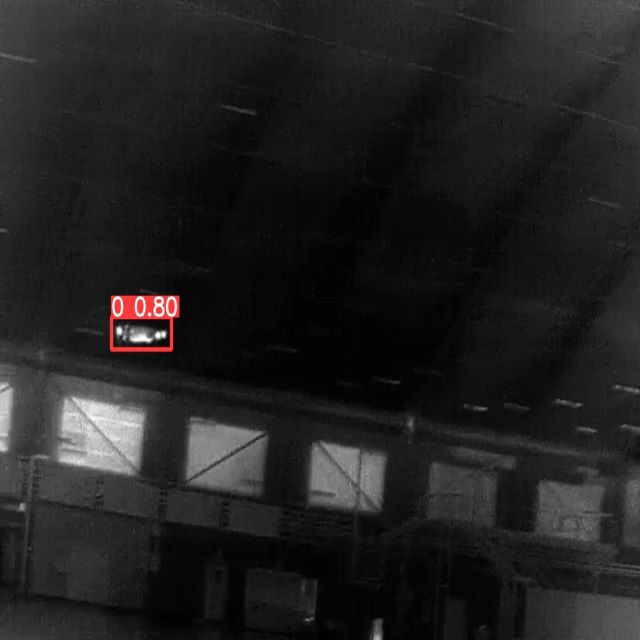}
         \includegraphics[width=\textwidth]{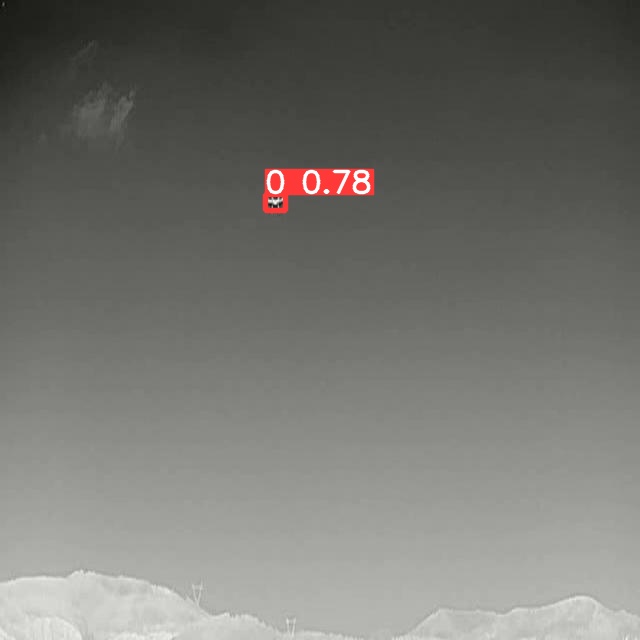}
         \includegraphics[width=\textwidth]{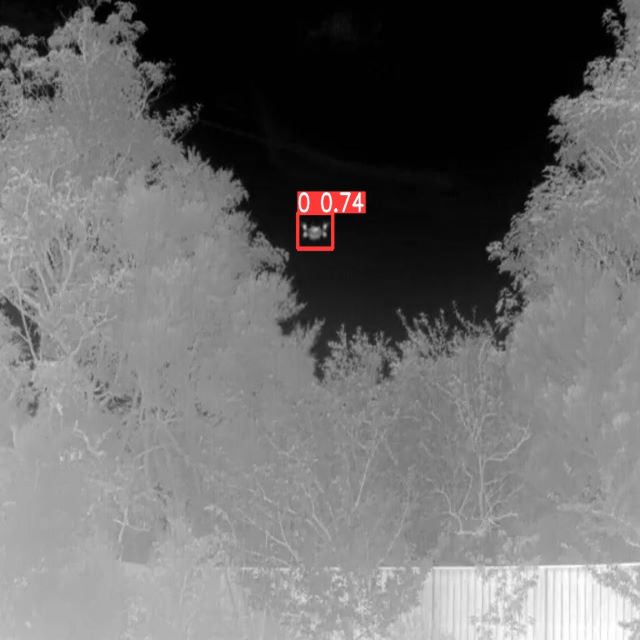}
         \includegraphics[width=\textwidth]{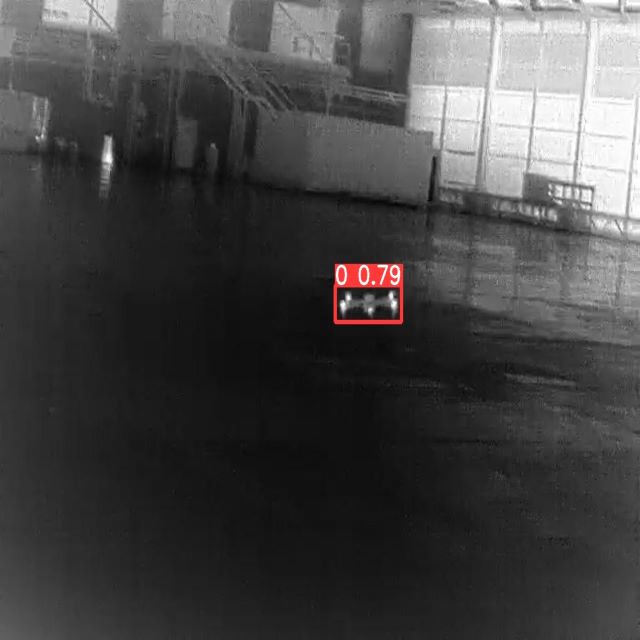}
         \caption{YOLOv5m}
         \label{figd}
     \end{subfigure}
     \hfill
     \begin{subfigure}[b]{0.09\textwidth}
         \centering
         \includegraphics[width=\textwidth]{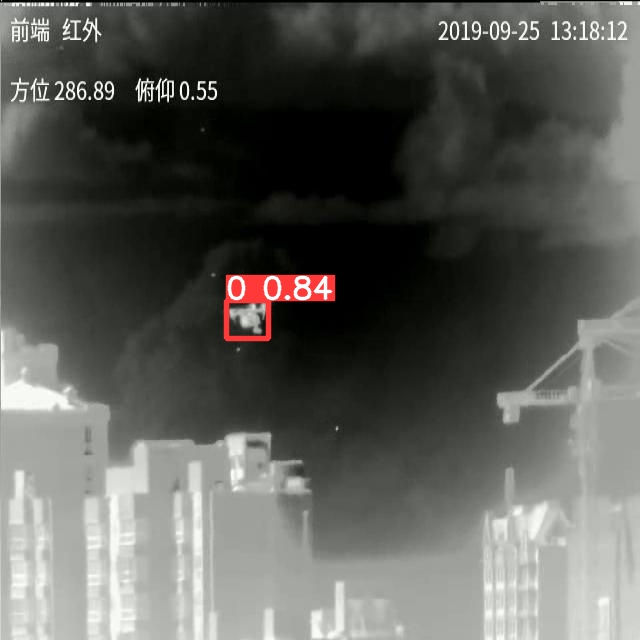}
         \includegraphics[width=\textwidth]{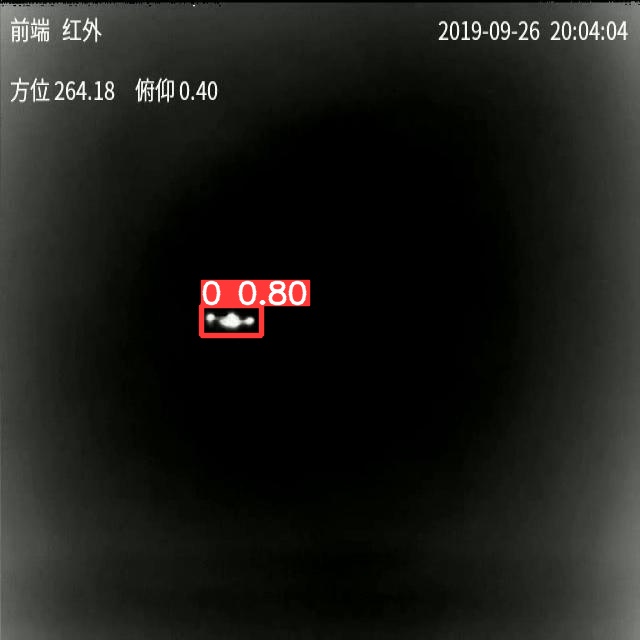}
         \includegraphics[width=\textwidth]{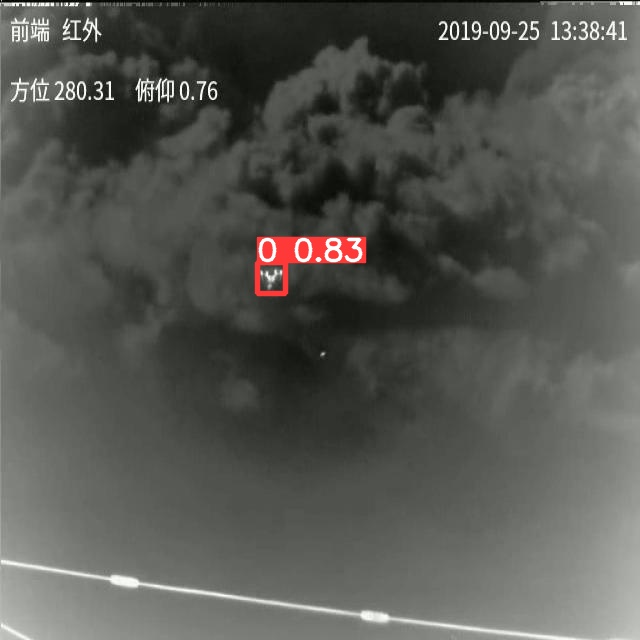}
         \includegraphics[width=\textwidth]{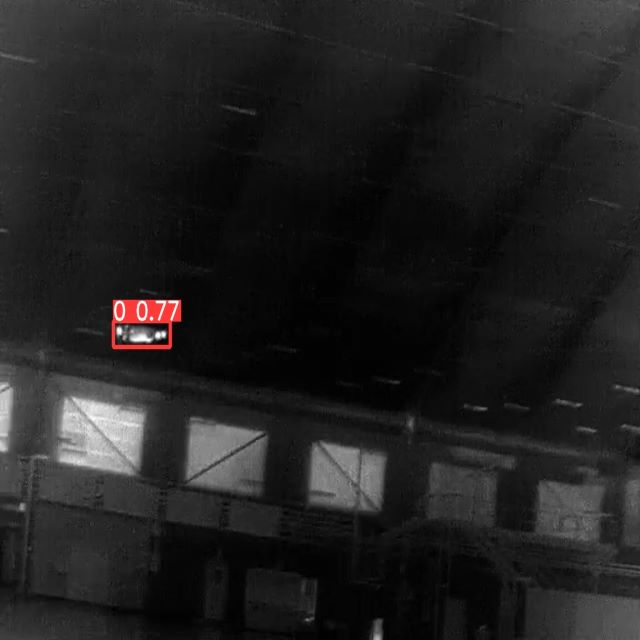}
         \includegraphics[width=\textwidth]{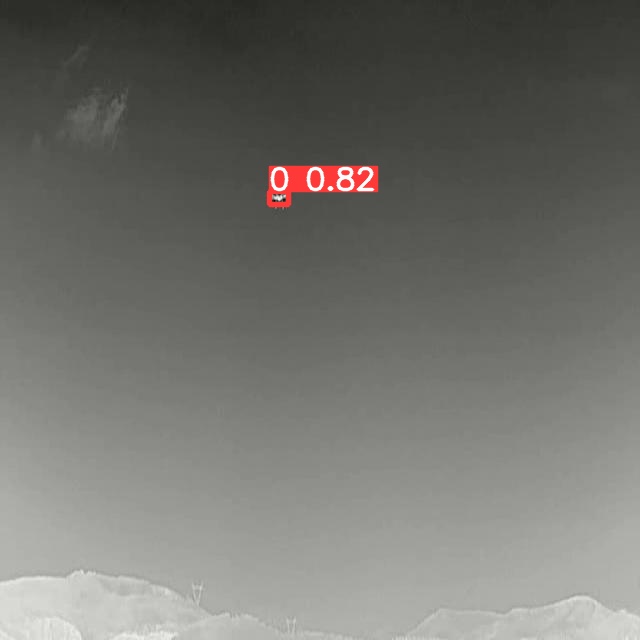}
         \includegraphics[width=\textwidth]{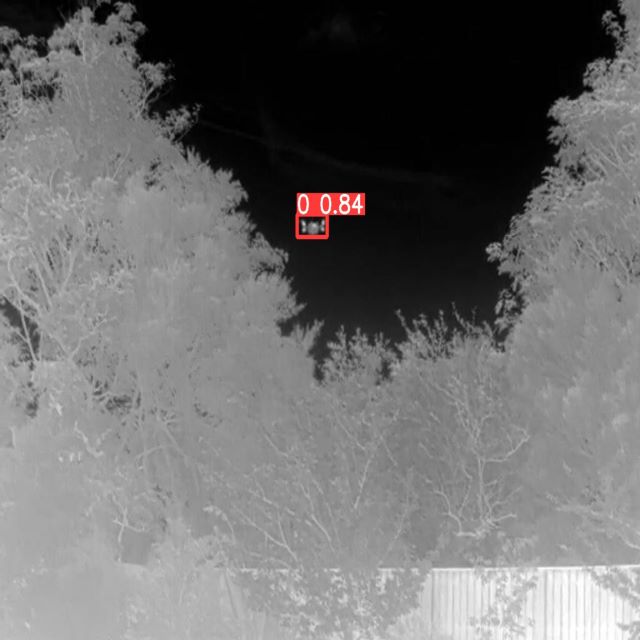}
         \includegraphics[width=\textwidth]{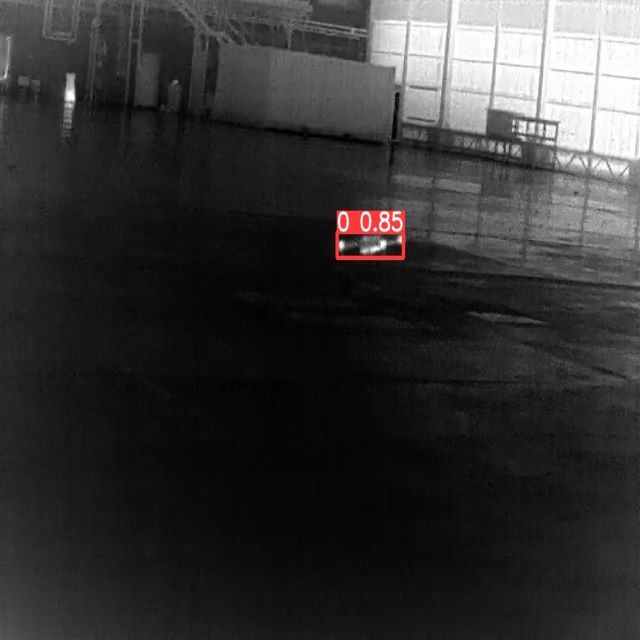}
         \caption{YOLOv5l}
         \label{fige}
     \end{subfigure}
     \hfill
     \caption{Night time UAV Detection with complex backgrounds (Top to bottom) like urban, clear sky, dense cloudy conditions, indoor settings, foggy weather, shrub and low altitude.}
        \label{f2}
\end{figure}
\subsection{Model Performance under Varying Environmental Backgrounds and Conditions}
To understand the impact of varying environmental backgrounds on detection performance, we compare the results of the proposed TF-Net with YOLOv5 models w.r.t different backgrounds including clearsky, urban, dense cloudy, indoor settings, foggy, shrub, and at low altitude in Fig. \ref{f2}. TF-Net achieved 93\% accuracy in the urban background compared to the YOLOv5s (89\% accuracy). Similarly, for dense cloudy and indoor settings, TF-Net performed well with 88\% and 91\%  accuracy, respectively. For foggy background, TF-Net yielded 4\% better accuracy than YOLOv5 models. TF-Net also performed better by achieving 13\% and 3\% higher accuracy in the case of shrubs and low-altitude UAVs, respectively. But TF-Net yielded better results in detecting UAVs at high altitudes compared to low altitudes. YOLOv5m achieves a detection accuracy of 87\% in urban, 88\% in a clear sky, and 83\% in dense cloudy conditions, which is comparatively less than the TF-Net. Similarly, YOLOv5l achieved 9\%, 10\%, and 5\% less detection accuracy compared to TF-Net in an urban, clear sky and dense cloudy situation, respectively. The worst performance is shown by YOLOv5n with a detection accuracy of 77\% in dense cloudy conditions, 70\% in indoor, and  75\% in foggy weather.  For low-altitude UAV detection, YOLOv5m remained the least sensitive with a detection accuracy of 79\%. Here, TF-Net is proved to be a better choice for UAV detection in all backgrounds and altitudes.

\subsection{Model Performance for Multi-size Target}
The modified kernel size enables the proposed TF-Net network to perform well on multi-size images. Fig. \ref{f3} shows multi-size UAV detection by TF-Net. Despite being lightweight, TF-Net efficiently detected small, medium, and large UAVs with a detection accuracy of 90\%, 91\%, and 91\%, respectively. YOLOv5s achieved 3\%, 4\%, and 2\% less detection accuracy on small, medium and, large UAVs, respectively compared to TF-Net. Among all four YOLOv5 models, YOLOv5s achieved higher accuracy's for multiple scale targets. The worst detection accuracy of 77\% is shown by YOLOv5l on medium sized drones . Highest accuracy's are shown by TF-Net on multiple sized UAV targets. This indicates the effectiveness and sensitivity of the proposed TF-Net model to detect UAVs of any target size in complex backgrounds.
\begin{figure}[h]
     \centering
     \begin{subfigure}[b]{0.1\textwidth}
         \centering
         \includegraphics[width=\textwidth]{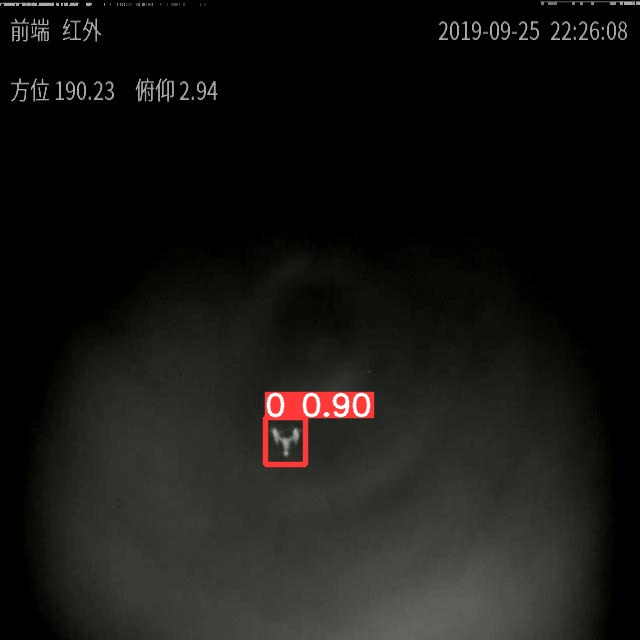}
         \includegraphics[width=\textwidth]{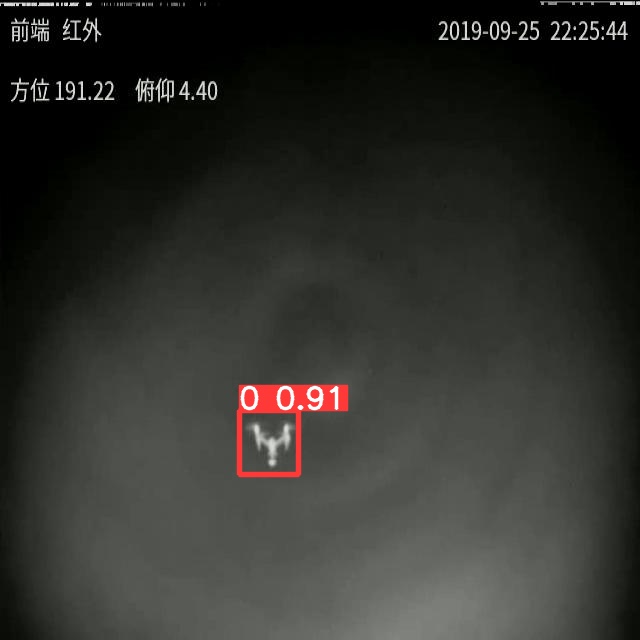}
         \includegraphics[width=\textwidth]{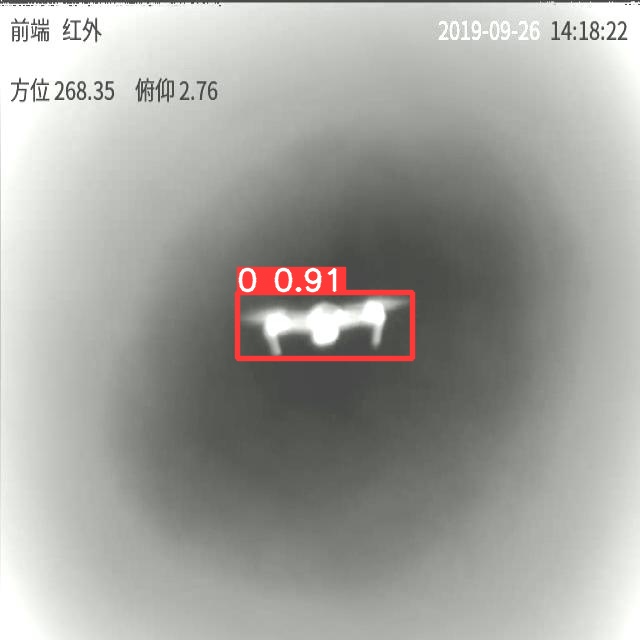}
         \caption{TF-Net}
         \label{figa}
     \end{subfigure}
     \hfill
     \begin{subfigure}[b]{0.1\textwidth}
         \centering
         \includegraphics[width=\textwidth]{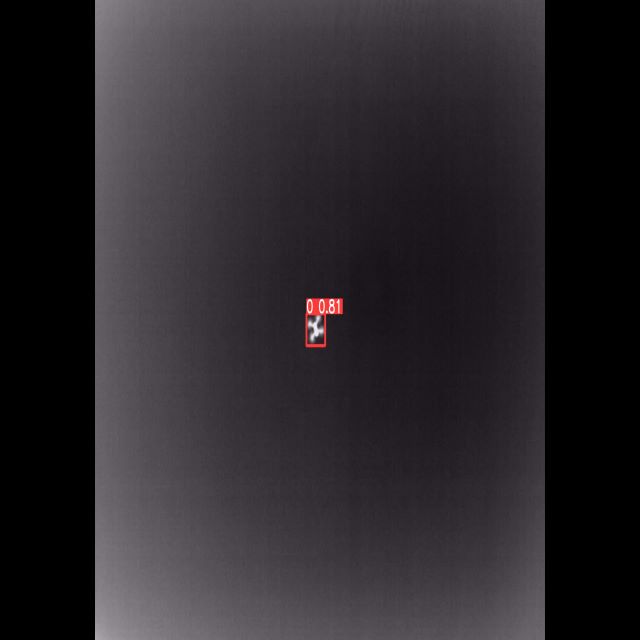}
         \includegraphics[width=\textwidth]{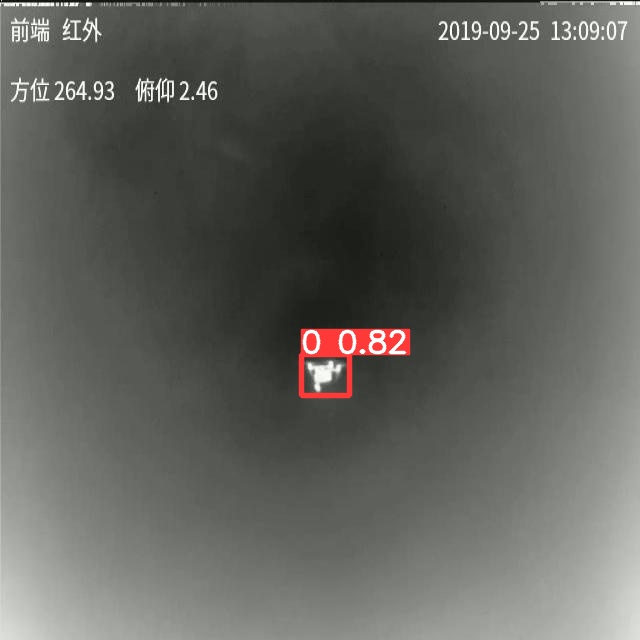}
         \includegraphics[width=\textwidth]{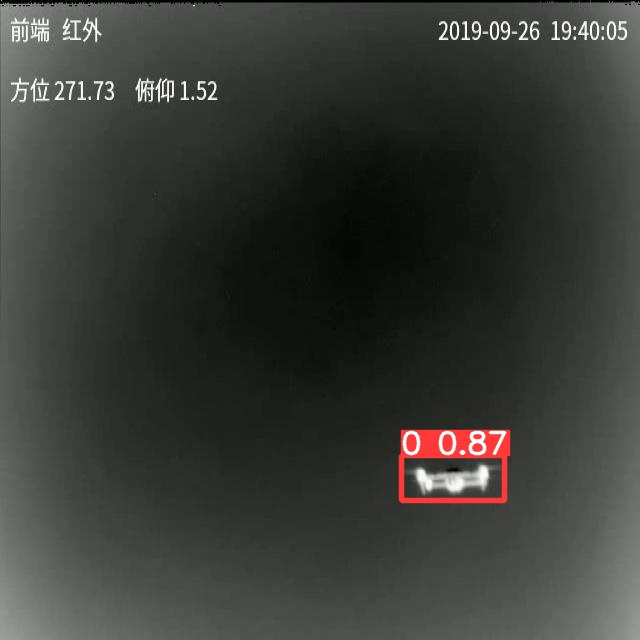}
         \caption{YOLOv5n}
         \label{figb}
     \end{subfigure}
     \hfill
     \begin{subfigure}[b]{0.1\textwidth}
         \centering
          \includegraphics[width=\textwidth]{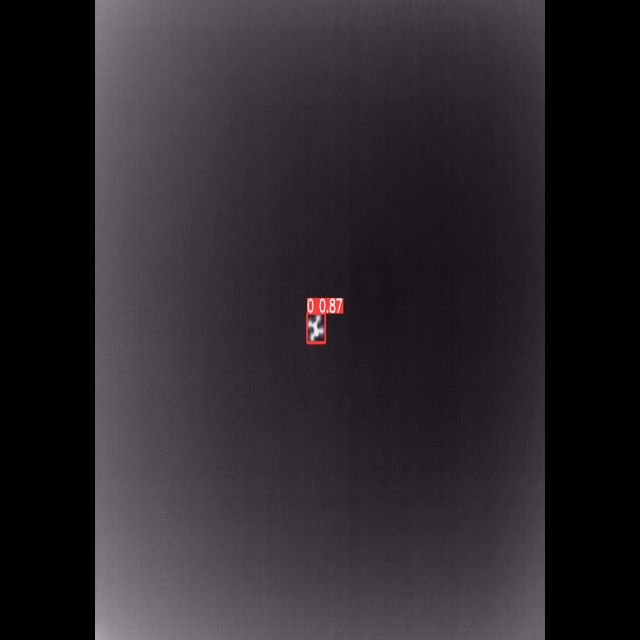}
         \includegraphics[width=\textwidth]{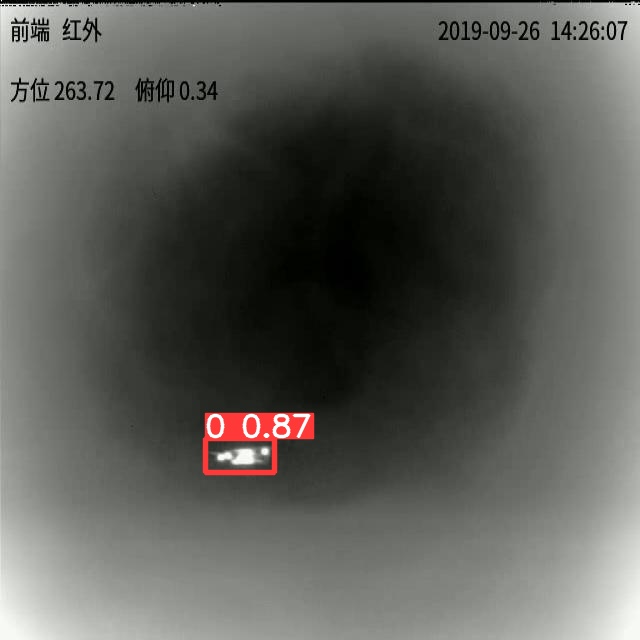}
         \includegraphics[width=\textwidth]{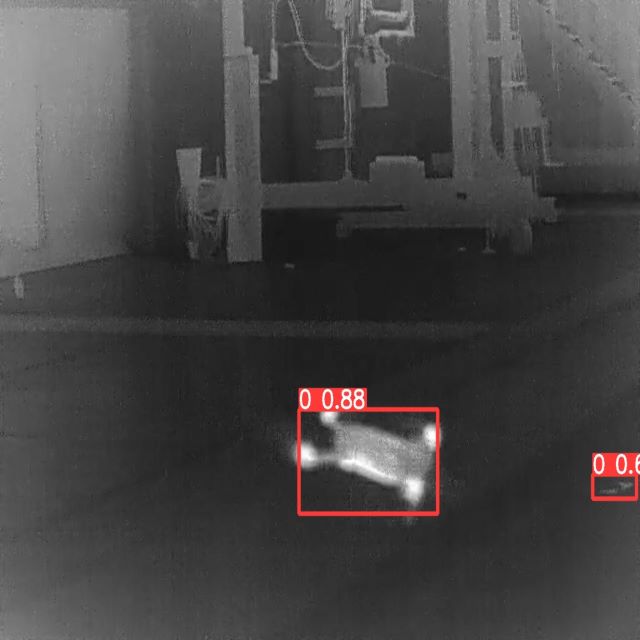}
         \caption{YOLOv5s}
         \label{figc}
     \end{subfigure}
     \hfill
     \begin{subfigure}[b]{0.1\textwidth}
         \centering
         \includegraphics[width=\textwidth]{5n_small.jpg}
         \includegraphics[width=\textwidth]{5n_med.jpg}
         \includegraphics[width=\textwidth]{5n_large.jpg}
         \caption{YOLOv5m}
         \label{figd}
     \end{subfigure}
     \hfill
     \begin{subfigure}[b]{0.1\textwidth}
         \centering
          \includegraphics[width=\textwidth]{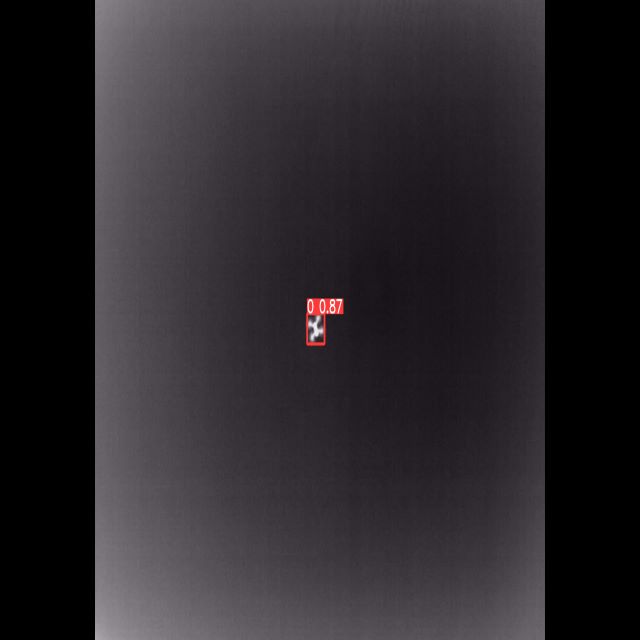}
         \includegraphics[width=\textwidth]{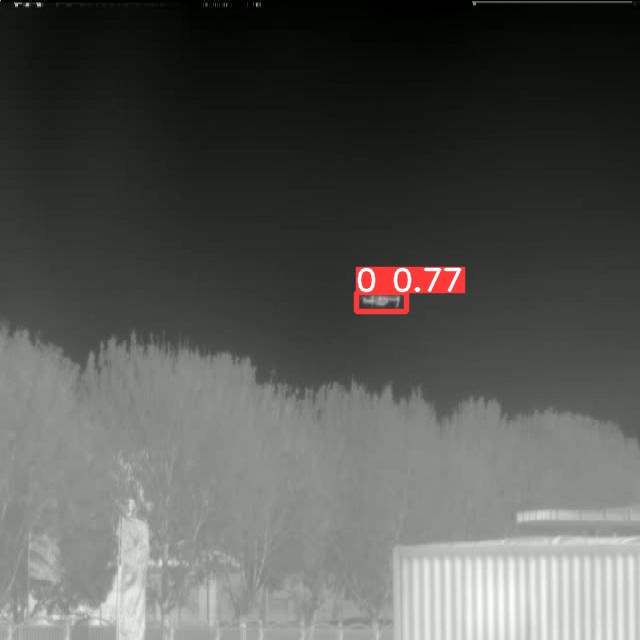}
         \includegraphics[width=\textwidth]{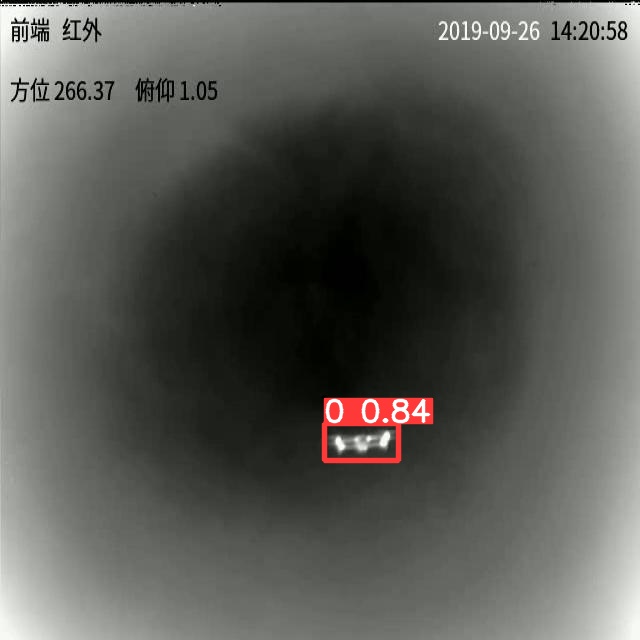}
         \caption{YOLOv5l}
         \label{fige}
     \end{subfigure}
     \hfill
     \caption{Detection results of Multi-size IR AV targets (Top to bottom) small, medium and large.}
        \label{f3}
\end{figure}

\subsection{Models Computational Complexity}
The system's computational complexity depends on training epochs plus time, trained model size with layers, and GFLOPs. For real-time object detection, it is important to have a lightweight model. In our case, the model size decreased to 10.8 MB, making it suitable and efficient for real-time UAV detection. For YOLOv5s, YOLOv5m, and YOLOv5l, the model size was 14.8MB, 42.1 MB, and 92.8 MB. This depicts a 4\% decrease in TF-Net model size compared to the YOLOv5s. TF-Net was trained for 292 epochs using 232 layers and extracted 5222070 parameters with 75.2 GFLOPs. TF-Net has higher GFLOPs compared to YOLOv5s (16.7 GFLOPs) despite having the same number of network layers.  The remaining training parameters are listed in Table \ref{T2}. The concept of early stopping was employed here and 300 epochs were given as input epoch to all models. Table \ref{T3} shows that each models stopped at different epoch. YOLOv5l stopped at the earliest epoch of 186 and has the highest GFLOPs of 107.6 while TF-Net stopped at 292 epoch. YOLOv5n took the least training time of 1.2 hours with a trained model size of 3.8 MB. Thus, TF-Net has the most comparable computational complexity with good GFLOPs performance (75.2), low trained model size (10.8MB) plus layers (232) and high precision (95.7\%).
\begin{table}[h]
\caption{Computational Complexity}
    \centering
    \begin{tabular}{p{2cm} p{2cm} p{2cm} p{2cm} p{2cm} p{2cm} }
    \hline
        \textbf{Model} &  \textbf{Epochs} &\textbf{Training Time (Hrs)} & \textbf{Size (MB)} & \textbf{Layers}  &  \textbf{GFLOPs} \\ \hline
        
        \textbf{TF-Net }& \textbf{292} & \textbf{4.5} & \textbf{10.8} & \textbf{232 } & \textbf{75.2} \\ \hline
      
        YOLOv5n & 200 ($\downarrow$ 92) & 1.2 ($\downarrow$ 3.3) & 3.8 ($\downarrow$ 7) & 213 ($\downarrow$ 19) & 4.1 ($\downarrow$ 71.1 ) \\ \hline
        YOLOv5s & 230 ($\downarrow$ 62) & 1.6($\downarrow$ 2.9 ) & 14.8 ($\uparrow$ 4) & 232 (same) & 16.7 ($\downarrow$ 58.5 ) \\ \hline
        YOLOv5m & 202 ($\downarrow$ 90) & 2.4 ($\downarrow$ 2.1) & 42.1 ($\uparrow$ 31.3) & 290 ($\uparrow$ 58)& 47.9 ($\downarrow$ 27.3)  \\ \hline
        YOLOv5l & 186 ($\downarrow$ 106)& 3.1 ($\downarrow$ 1.4) & 92.8 ($\uparrow$ 82) & 367 ($\uparrow$ 135 ) & 107.6 ($\uparrow$ 32.4 ) \\ \hline
    \end{tabular}
    \label{T3}
\end{table}

\subsection{Inferences time and Speed}
The improvement of the UAV detection speed and real-time system implementation rely greatly on the inference rate. We were able to reach a detection time of 10.2 ms due to the features' computational interdependence properties. TF-Net achieved 0.7 ms NMS per image with 98 FPS trained on Tesla T4 GPU mentioned in Table \ref{T2}. This shows the effectiveness of TF-Net model in real time scenarios for fast UAV detection. 
\begin{table}
\caption{Model Evaluation on Test Data}
    \centering
    \begin{tabular}{l l l l l }
    \hline
      \textbf{Model} & \textbf{Pre-process (ms)} & \textbf{Inference(ms)}&	\textbf{NMS per image (ms)}	& \textbf{FPS}
 \\ \hline
       \textbf{TF-Net} & \textbf{0.3}  &	\textbf{10.2}	& \textbf{0.7}	& \textbf{98} \\ \hline
        \textbf{YOLOv5n} & 0.3 (same)	& 8.4 ($\downarrow$ 1.8)	& 0.9 ($\uparrow$ 0.2)	& 119 ($\uparrow$ 21)  \\ \hline
        \textbf{YOLOv5s} & 0.3(same)	&	9.3 ($\uparrow$ 0.2)	& 0.7(same)	& 107 ($\uparrow$ 9)   \\ \hline
        \textbf{YOLOv5m} & 0.4 ($\uparrow 0.1$)&	14 ($\uparrow$ 3.8)	&0.9 ($\uparrow$ 0.2)	&71  ($\downarrow$ 27)\\ \hline
        \textbf{YOLOv5l} & 0.3(same)	&15.4 ($\uparrow$ 5.2)	&0.7(same)	&64 ($\downarrow$ 34) \\ \hline
    \end{tabular}
    \label{T2}
\end{table}
\subsection{Comparison With the State-of-the-Art}
We compare our results with the state-of-the-art models for UAV detection based on YOLOv4 and YOLOv5 models. It is apparent from Table \ref{T4} that TF-Net performed well in terms of precision, FPS, and model size compared to the other literature models. TF-Net outperformed all other models and achieved 95.7\% precision. When compared to \cite{5}, \cite{20} and \cite{8}, TF-Net has the highest FPS of 98. TF-Net also proved to be light-weight with a trained model size of 10 MB and 5.2 million parameters as compared to \cite{23}. Therefore we implemented night vision UAV detection utilizing the proposed TF-Net model with IR images and achieved superior results compared to the literature and baseline models.
\begin{table}
    \caption{Comparison With the State-of-the-Art}
    \begin{tabular}{p{2cm} p{2cm} p{2cm} p{1.5cm} p{1.5cm} p{2cm} p{2cm} p{2cm}}
    \hline
        \textbf{Model} & \textbf{Dataset (images)}& \textbf{Input size} &  \textbf{Precision  (\%)} & \textbf{FPS} & \textbf{Parameters (Millions)} & \textbf{CPU/GPU}  \\ \hline
        \textbf{TF-Net }& \textbf{3891} & 416$\times$416 & \textbf{95.7} & \textbf{98 } & \textbf{5.2 } &   \textbf{NVIDIA Tesla T4}\\ \hline
      
        \textbf{Improved YOLOv5 \cite{9}} & 1259 & 640$\times$640 & 94.96  & N/G & 7.02 &NVIDIA GeForce RTX 3050 \\ \hline
        \textbf{Pruned YOLOv4 \cite{10}} & 10,000 & 416$\times$416 & 74.2 & 43 & 63.9 & Tesla P100  \\ \hline
        \textbf{Real-time Drone Detection Algorithm \cite{15}} & 24,075 & N/G & 70.1 & 9.5 & N/G & NVIDIA GeForce GT 1030\\ \hline
        \textbf{YOLOv4 \cite{12}} & 4032 & 416$\times$416 & 89.32 & 38 & N/G & N/G \\ \hline  
        \textbf{YOLOv5 \cite{17}} & 1359  & 416$\times$416 & 94.7 & N/G & 87 & NVidia RTX2070\\ \hline  
    \end{tabular}
    \label{T4}
\end{table}

 \section{Conclusion}
We proposed an improved version of the YOLOv55s based object detection model TF-Net to improve UAV detection in adverse weather conditions at night, where each input image was adaptively enhanced to obtain better detection performance. Kernel-based feature-map extraction was used to extract weather-specific target information for the TF-Net detector, whose hyper-parameters are set to attain excellently optimized results. Moreover, the whole framework was trained in an end-to-end fashion, where the parameter prediction network was weakly supervised to learn small sized feature maps with less detection loss. By taking advantage of excellent training, early screening, and feature map based prediction, our proposed approach was able to adaptively handle normal and adverse weather conditions. The experimental results showed that our method performed much better than previous approaches in both the foggy and low-light scenarios. The proposed TF-Net model performed well in terms of precision, model size, mAp@0.5, $IoU$ and GFLOPs compared to the baseline YOLOv5 models. The TF-Net achieved 4.5\% higher precision, compared to the baseline YOLOv5 model.
%
%
%
\bibliographystyle{splncs04}
\bibliography{TFNET}

\end{document}